\documentclass[11pt, a4paper, logo, copyright, nonumbering]{xiaomi}

\usepackage[numbers, sort&compress, square]{natbib}

\usepackage{amsmath,amsfonts,bm}

\def\eqref#1{equation~\ref{#1}}
\def\Eqref#1{Equation~\ref{#1}}

\def\1{\bm{1}}

\def\vmu{{\bm{\mu}}}

\def\va{{\bm{a}}}

\def\vg{{\bm{g}}}

\def\vv{{\bm{v}}}

\def\vz{{\bm{z}}}

\def\eva{{a}}

\def\mP{{\bm{P}}}

\DeclareMathAlphabet{\mathsfit}{\encodingdefault}{\sfdefault}{m}{sl}
\SetMathAlphabet{\mathsfit}{bold}{\encodingdefault}{\sfdefault}{bx}{n}
\newcommand{\tens}[1]{\bm{\mathsfit{#1}}}
\def\tA{{\tens{A}}}
\def\tB{{\tens{B}}}

\def\tH{{\tens{H}}}

\def\tM{{\tens{M}}}

\def\tR{{\tens{R}}}

\def\tU{{\tens{U}}}
\def\tV{{\tens{V}}}

\def\tX{{\tens{X}}}
\def\tY{{\tens{Y}}}
\def\tZ{{\tens{Z}}}

\def\sR{{\mathbb{R}}}

\usepackage{hyperref}
\usepackage{url}
\usepackage{tcolorbox}
\tcbuselibrary{breakable,listings}
\usepackage{listings}
\usepackage{subcaption}
\usepackage{enumitem}
\usepackage{wrapfig}
\usepackage{capt-of}
\usepackage{cuted}
\usepackage{array}
\usepackage{tabularx}
\usepackage{xcolor}
\usepackage{colortbl}
\usepackage{marvosym}

\usepackage[capitalize,nameinlink]{cleveref}

\usepackage{booktabs,graphicx,amsmath}
\usepackage{natbib}

\newcommand{\teaserfigure}{%
    \begin{center}
        \includegraphics[width=\linewidth]{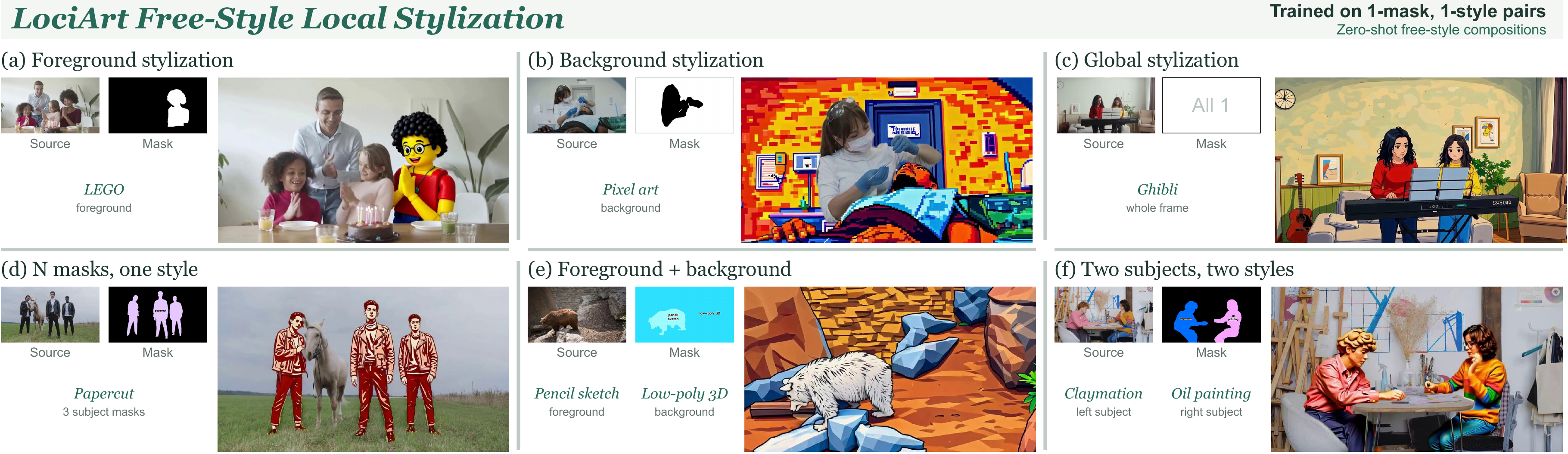}
        \captionof{figure}{\textbf{Examples of free-style local video stylization.}
        LociArt supports local or global editing, shared or distinct styles
        across multiple regions, and style composition without compositional
        training pairs. Small images show the source videos and masks;
        white and black denote editing and preservation, respectively.
        Color masks identify styles.}
        \label{fig:teaser}
    \end{center}
    \vspace{1em}
}

\makeatletter
\let\xiaomiabscontent\abscontent
\renewcommand{\abscontent}{%
    \teaserfigure
    \xiaomiabscontent
}
\makeatother

\definecolor{DeltaGreen}{HTML}{218739}
\definecolor{DeltaRed}{HTML}{C23A36}
\newcommand{\oldgain}[1]{\,{\scriptsize\textcolor{DeltaGreen}{\ensuremath{+\Delta#1}}}}
\newcommand{\olddrop}[1]{\,{\scriptsize\textcolor{DeltaRed}{\ensuremath{-\Delta#1}}}}

\title{LociArt: A Unified Paradigm for Free-Style Local Video Stylization}

\author{
Yuxuan Chen$^{1*}$ \quad
Feng Xie$^{2*}$ \quad
Jiagao Hu$^{1*}$\textsuperscript{\Letter} \quad
Fuhao Li$^{1}$ \quad
Zepeng Wang$^{1}$ \quad
Fei Wang$^{1}$ \quad
Daiguo Zhou$^{1}$
}
\institute{
$^{1}$ MiLM Plus, Xiaomi Inc.\\
$^{2}$ Xi'an University of Electronic Science and Technology\\
{\footnotesize\textsuperscript{*}The first three authors contributed equally.}
}

\begin{document}

\begin{abstract}
Video stylization typically transforms entire scenes, while local artistic editing requires precise region--style association, spatial control, and temporal consistency.
We introduce \textbf{LociArt}, a unified paradigm for free-style local video stylization that explicitly specifies \emph{where to stylize and what style to apply} while preserving non-target content.
At its core, \textbf{GRNStyle} builds on Generative Refinement Networks (GRN) and introduces a polar style-aware mask that embeds style identities as equal-norm directions within selected regions.
Complementary representation--geometry conditioning combines encoded mask features with raw-mask spatial support, integrating spatiotemporal context with geometric control.
To provide paired supervision, we develop an automated pipeline combining global stylization, video object removal, matting, and compositing, and construct the \textbf{LociArt-100K Dataset} with 100K local stylization pairs.
We further establish the \textbf{LociArt Benchmark}, comprising 80 source videos and 84 target masks covering 168 foreground/background stylization cases, with a \textbf{Style Prototype Similarity (SPS)} evaluation scheme for regional style quality and complementary measures of non-target preservation and temporal consistency.
Experiments show that GRNStyle outperforms existing methods in regional style quality and spatial control while maintaining strong temporal consistency.
Qualitative examples further illustrate joint stylization of multiple regions with the same or different learned styles, despite training only on single-region, single-style pairs.
Together, these components provide a unified foundation for learning and evaluating controllable local video stylization.
Project page: \url{https://higherhu.github.io/lociart/}.
\end{abstract}

\maketitle

\section{Introduction}
\label{sec:intro}

Video stylization transforms recorded footage into artistic renditions for film, advertising, and digital art. Building on image style transfer \citep{gatys2016image,huang2017arbitrary} and diffusion-based image generation and editing \citep{rombach2022high,brooks2023instructpix2pix}, video methods use text or reference images to guide appearance while preserving motion \citep{geyer2024tokenflow,ye2025stylemaster,song2026vista}. Yet artistic control often requires more than a global transformation: an artist may turn a moving subject into a sketch within an otherwise unchanged scene, reimagine the surroundings as a watercolor while keeping the subject intact, or bring distinct artistic styles to different objects in the same shot, as shown in \cref{fig:teaser}. We call this task \emph{free-style local video stylization}: controlling both \emph{where to stylize} and \emph{what style to apply} across space and time, while preserving non-target content.

Video segmentation provides a practical starting point. A few user interactions can define masks that propagate across frames, with streaming inference supporting real-time processing \citep{ravi2025sam,carion2026sam}. Such masks offer an accessible spatial prior for local stylization \citep{song2025univst,he2026stylegallery}, extending control from foreground subjects to the less-explored task of stylizing their surroundings. Yet selecting a region does not specify its artistic transformation: a binary mask indicates editing extent but not style identity. With style supplied separately through text, the model must learn to associate the intended appearance with the selected region.

Reliable local control also requires preserving region boundaries throughout motion. Despite advances in attention supervision \citep{chen2026regionroute,long2026coinve} and mask-conditioned editing \citep{jiang2025vace,bian2025videopainter, li2026efficient, kang2026mast}, mask representations face a trade-off between spatiotemporal context and spatial fidelity. VAE-based processing aligns masks with learned video features \citep{yatim2025dynvfx}, but compression can blur their geometry; direct resampling retains spatial structure without the same learned temporal context~\citep{hu2026ideal}. Thus, a useful conditioning mechanism must both bind styles explicitly to regions and retain the spatial detail needed to localize their effects over time.

These modeling challenges are compounded by gaps in supervision and evaluation. Training pairs must demonstrate local artistic changes, faithful preservation elsewhere, and natural transitions along moving boundaries. Global stylization pairs \citep{song2026vista,li2026dreamstyle} do not directly supply this supervision, while instruction-based video-editing datasets \citep{bai2026scaling,long2026coinve} cover broader editing operations. Likewise, global similarity and video-averaged scores can conceal weak regional style, changes outside the target, and transient failures. Free-style local video stylization therefore calls for coordinated advances in paired data, region--style conditioning, and region-aware evaluation.

We introduce \emph{LociArt}, a unified paradigm addressing these needs through the \textbf{LociArt-100K Dataset}, the \textbf{GRNStyle} model, and the \textbf{LociArt Benchmark}. Our automated data pipeline combines global stylization, video object removal, matting, and compositing to construct 100K curated pairs. Each example contains a source video, an editing mask, a style instruction, and a locally stylized target, supervising transformation and preservation together. Constructing these targets requires handling contour changes introduced by stylization. Object removal supplies clean background content for newly exposed areas, while alpha mattes estimated separately for original and stylized objects guide compositing along their respective contours, avoiding object remnants and enabling natural boundary transitions.

Trained on these pairs, \textbf{GRNStyle} builds on Generative Refinement Networks (GRN) \citep{han2026grn} and the source-evidence editing formulation of GRNEdit \citep{xie2026grnedit}. Its core design is a \emph{polar style-aware mask}, which embeds style identities as equal-norm directions within selected regions, unifying \emph{where to stylize} and \emph{what style to apply} without imposing a scalar ordering on styles. Complementary \emph{representation--geometry conditioning} combines VAE-encoded mask features with raw-mask spatial support to integrate spatiotemporal context and geometric control, while source evidence guides content preservation throughout refinement.

To evaluate local stylization, the \textbf{LociArt Benchmark} contains 80 source videos and 84 object--mask units, each evaluated under both foreground and background stylization protocols, yielding 168 stylization cases. It uses \emph{Style Prototype Similarity (SPS)} to measure regional style fidelity, together with complementary measures of non-target preservation and temporal consistency.

Our contributions are threefold:
\begin{itemize}[leftmargin=1em, nosep]
    \item \textbf{A Unified Paradigm for Local Video Stylization.} We introduce LociArt, a unified computational paradigm that formulates local video stylization as region-conditioned artistic transformation, explicitly specifying where to stylize and what style to apply while preserving non-target content and temporal consistency.
    \item \textbf{GRNStyle with Style-Aware Region Control.} We propose GRNStyle, a GRN-based model that unifies region and style through polar style-aware masks and combines spatiotemporal context with geometric control. Qualitative examples further illustrate free-style multi-region, multi-style compositions learned from single-region, single-style supervision.
    \item \textbf{Scalable Data Construction and Region-Aware Evaluation.} We develop an automated pipeline to construct the LociArt-100K Dataset and establish the LociArt Benchmark with 80 source videos and 84 object--mask units, integrating SPS with preservation and temporal consistency measures for region-aware evaluation.
\end{itemize}

\section{Related Work}
\label{sec:related}

\noindent\textbf{Image style transfer.}
Neural style transfer progressed from per-image optimization~\citep{gatys2016image} to feed-forward arbitrary transfer with AdaIN~\citep{huang2017arbitrary}.
Building on latent diffusion~\citep{rombach2022high}, StyleShot~\citep{gao2025styleshot} separates style from content, while TeleStyle~\citep{zhang2026telestyle} learns content-preserving transfer from curated triplets and extends it to video.
Localized editing, however, requires explicit spatial control.

\noindent\textbf{Localized stylization.}
Image methods use masked feature or attention control~\citep{li2026efficient}, mask-learned regional attention~\citep{chen2026regionroute}, or semantic-region matching~\citep{he2026stylegallery}.
For video, Xia et al.\cite{xia2021real} use spatiotemporal alignment, while UniVST~\citep{song2025univst} combines mask propagation, AdaIN-guided latent and attention manipulation, and optical-flow smoothing without training.
In contrast, we study locally artistic video stylization with paired supervision and fixed, equal-norm polar codes that bind style identity to editable regions.

\noindent\textbf{Video stylization and generative editing.}
PickStyle~\citep{mehraban2025pickstyle} learns lightweight style adapters from synthetic clips derived from paired images.
VISTA~\citep{song2026vista} and MoStyle~\citep{wei2026stylized} use style/source/stylized-video triplets; DreamStyle~\citep{li2026dreamstyle} supports text-, image-, and first-frame-guided stylization from paired videos.
For broader editing, VACE~\citep{jiang2025vace} unifies video conditions including masks, while VideoPainter~\citep{bian2025videopainter} performs context-conditioned inpainting.
Beyond diffusion, GRN~\citep{han2026grn} uses binary-state refinement, and GRNEdit~\citep{xie2026grnedit} treats the source as editing evidence.
GRNStyle extends this evidence-based formulation by combining VAE-encoded mask features with raw-mask geometry, thereby coupling temporal semantics with spatial fidelity.

\noindent\textbf{Data and evaluation.}
Ditto-1M~\citep{bai2026scaling} scales instruction-based editing, while CoinVE-200K and CoinVE-Bench~\citep{long2026coinve} target compositional editing, including regional stylization.
StyleShot~\citep{gao2025styleshot} and MegaStyle~\citep{gao2026megastyle} provide image data for stylistic consistency and diversity; StyleShot also introduces StyleBench.
Evaluation spans image quality with ArtFID~\citep{wright2022artfid}, style similarity with CSD~\citep{somepalli2024measuring}, and video quality and temporal consistency with VBench~\citep{huang2024vbench}; RegionRoute~\citep{chen2026regionroute} also evaluates regional style and background preservation.
LociArt-100K adds pairs aligning content, regions, styles, and targets; its benchmark measures regional style, non-target preservation, temporal consistency, and VLM-based perceptual quality.

\begin{figure*}[t]
    \centering
    \includegraphics[width=\linewidth]{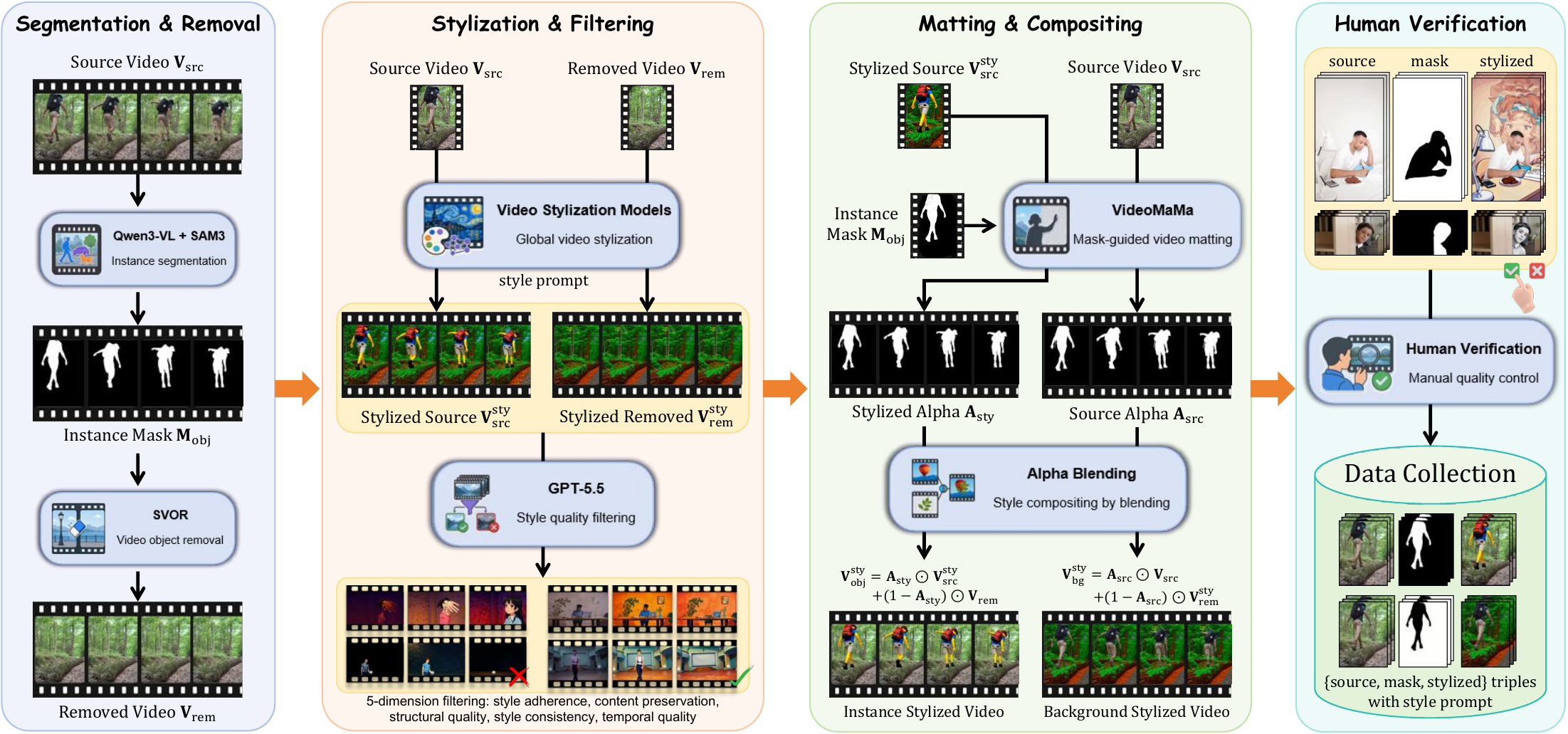}
    \caption{\textbf{Overview of the data construction pipeline.} Source and object-removed videos are globally stylized and filtered, then combined through mask-guided matting to produce foreground- and background-stylized pairs for human verification.}
    \label{fig:data_pipeline}
\end{figure*}

\section{LociArt-100K Dataset Construction}
\label{sec:data_pipeline}

To supervise local stylization and non-target preservation, we construct the \textbf{LociArt-100K Dataset} of 100K pairs by separating global style generation from spatial composition. Specifically, we globally stylize source and object-removed videos, then recombine their components into complementary foreground and background targets. Each example contains a source video $\tV_{\mathrm{src}}$, binary editing mask $\tM$, style instruction $p$, and locally stylized target $\tV^{\mathrm{sty}}$.
As illustrated in~\cref{fig:data_pipeline}, the pipeline comprises the following four steps.

\paragraph{Step 1: Object identification and removal.}
We first identify the objects to separate from their surroundings. Source videos are collected from SA-V~\citep{ravi2025sam} and Ditto-1M~\citep{bai2026scaling}. Qwen3-VL-32B~\citep{bai2025qwen3} identifies salient objects, and their category prompts guide SAM3~\citep{carion2026sam} to produce temporally tracked masks $\tM_{\mathrm{obj}}$. After vision-language, temporal-coverage, and area checks, each accepted mask guides SVOR~\citep{hu2026ideal} to remove the selected object, yielding $\tV_{\mathrm{rem}}$. This reconstructed background fills areas exposed by stylization-induced contour changes, avoiding original-object remnants during compositing. Appendix~\cref{sec:stage1_details} details the mask selection and filtering procedures.

\paragraph{Step 2: Global stylization and filtering.}
With source and reconstructed background videos available, we run both Ditto~\citep{bai2026scaling} and our internal fine-tuned VACE~\citep{jiang2025vace} on each under a stylization instruction $p$. For either global stylizer $G$, we get two stylized outputs:
\begin{equation}
    \tV_{\mathrm{src}}^{\mathrm{sty}} = G(\tV_{\mathrm{src}}, p),
    \qquad
    \tV_{\mathrm{rem}}^{\mathrm{sty}} = G(\tV_{\mathrm{rem}}, p).
\end{equation}
Before using these outputs for local composition, GPT-5.5 is used to assess style quality, content preservation, and temporal quality through two cascaded gates: the source-video pair is evaluated first, followed by the object-removed pair. Only candidates passing both proceed to matting and compositing. More details appear in Appendix~\cref{sec:stage2_details}.

\paragraph{Step 3: Matting and localized compositing.}
We next localize the accepted stylizations. Since stylization can change object contours, we use VideoMaMa~\citep{lim2026videomama} to estimate separate alpha mattes $\tA_{\mathrm{src}}$ and $\tA_{\mathrm{sty}}$ from $\tV_{\mathrm{src}}$ and $\tV_{\mathrm{src}}^{\mathrm{sty}}$ with the same tracked mask $\tM_{\mathrm{obj}}$, respectively. These mattes follow the respective object boundaries through time and are used for compositing instead of binary masks. The resulting composites stylize the foreground while retaining the background, or vice versa:
\begin{equation}
\begin{aligned}
    \tV_{\mathrm{fg}}^{\mathrm{sty}}
    &= \tA_{\mathrm{sty}} \odot \tV_{\mathrm{src}}^{\mathrm{sty}}
    + (1-\tA_{\mathrm{sty}}) \odot \tV_{\mathrm{rem}},\\
    \tV_{\mathrm{bg}}^{\mathrm{sty}}
    &= \tA_{\mathrm{src}} \odot \tV_{\mathrm{src}}
    + (1-\tA_{\mathrm{src}}) \odot \tV_{\mathrm{rem}}^{\mathrm{sty}}.
\end{aligned}
\label{eq:local_compositing}
\end{equation}
Here $\odot$ denotes element-wise multiplication, with alpha values broadcast across color channels. Both targets are paired with $\tV_{\mathrm{src}}$ and $p$. The directional editing mask is $\tM=\tM_{\mathrm{obj}}$ for the foreground target and $\tM=1-\tM_{\mathrm{obj}}$ for the background target.

\paragraph{Step 4: Human verification.}
Finally, because compositing may introduce artifacts, annotators inspect each completed example for content preservation, boundary quality, and temporal consistency. Only qualified samples are retained, yielding a final dataset of 100K video pairs covering 18 style categories.
More details on the verification and final dataset statistics can be found in \cref{sec:stage4_details}.

\begin{figure*}[t]
    \centering
    \includegraphics[width=\linewidth]{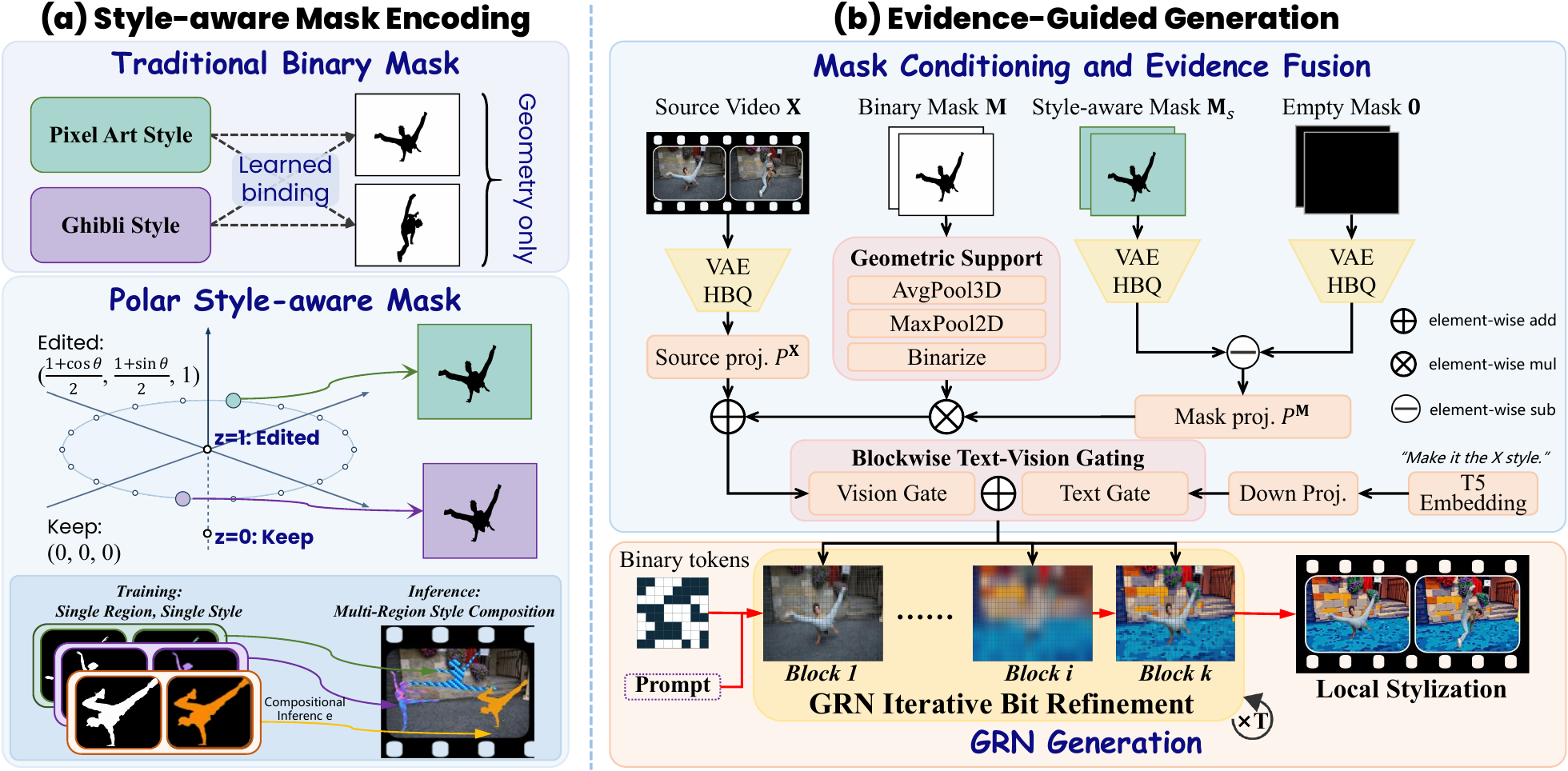}
    \caption{\textbf{Overview of GRNStyle.}
(a) Polar style-aware masks bind styles to regions and support multi-region composition.
(b) Anchored mask evidence and geometric support are combined with source features through block-wise gating to guide local video stylization.}
    \label{fig:model_pipeline}
\end{figure*}

\section{GRNStyle: Free-Style Local Video Stylization}
\label{sec:model_detail}

As the model within LociArt, \textbf{GRNStyle} binds style identities to target regions and combines encoded mask features with explicit geometric support (\cref{fig:model_pipeline}). Source evidence guides content preservation throughout generation, while the same spatial conditioning interface accommodates multiple regions at inference.

\subsection{Problem Formulation and Binary Evidence}\label{sec:problem}

Given a source video $\tX=\tV_{\mathrm{src}}\in[0,1]^{N\times H\times W\times3}$ with $N$ frames of spatial size $H\times W$, an editing mask $\tM\in\{0,1\}^{N\times H\times W\times1}$, and a text instruction $p$, we derive the target style $s\in\{1,\ldots,K\}$ from $p$, where $K$ is the number of style categories. We embed $s$ into $\tM$ to form a style-aware mask $\tM_s$ and learn $\widehat{\tY}=F(\tX,\tM_s,p\,;\Theta)$ from locally stylized targets $\tY=\tV^{\mathrm{sty}}$, aiming to transform the selected region while preserving non-target content.

To preserve source content during generation, we build on GRN~\citep{han2026grn} and GRNEdit's source-evidence formulation~\citep{xie2026grnedit}. An encoder $\mathcal E$ and binary quantizer $\mathcal Q$ map the source to $\tB^{\tX}=\mathcal Q(\mathcal E(\tX))\in\{0,1\}^{L\times D}$, with $L$ latent positions and $D$ bits per position. GRN uses one-hot bit codes $\phi(b)=(1-b,b)$ and parallel iterative refinement, while GRNEdit supplies source evidence for retaining or changing target bits. GRNStyle adds explicit style--region binding and geometric control to localize the transformation.

\subsection{Equal-Norm Style--Space Conditioning}\label{sec:polar}

A binary mask specifies where to edit but leaves the associated style to a separate text condition. To bind style directly to space, we assign each style a fixed polar angle $\theta_s$ at unit radius and embed its coordinates within the selected region (\cref{fig:model_pipeline}(a)). The first two channels of the three-channel VAE input store the rescaled Cartesian coordinates, i.e., $\va_s = (\cos\theta_s,\sin\theta_s)$, while the third retains binary occupancy for the polar style-aware mask:
\begin{equation}
\tM_s(q) = \tM(q)\left(\frac{1+\eva_{s,1}}2,\frac{1+\eva_{s,2}}2,1\right).
\label{eq:polar}
\end{equation}
Here $q$ indexes spatiotemporal pixel positions. Since $\|\va_s\|_2=1$, styles differ by direction rather than magnitude, avoiding an arbitrary scalar ordering. Unedited positions remain zero, and occupancy is directly recoverable from the third channel. This representation adds neither channels nor conditioning tokens relative to a repeated binary mask. Appendix~\cref{app:codebook_normalization} provides encoding details.

Crucially, the resulting field contains the region--style interaction before video encoding, as a linear readout $\mathcal R:\mathbb R^3\rightarrow\mathbb R^2$ recovers
\begin{equation}
\mathcal R(\tM_s(q))=\tM(q)\va_s.
\label{eq:readout}
\end{equation}
Thus, the input explicitly associates each edited position with its style code. Appendix~\cref{app:readout} derives this readout.

\subsection{Representation--Geometry Mask Conditioning}\label{sec:geometry}

The mask must be encoded on the backbone's latent grid during generation. However, common mask compression strategies can compromise spatial localization~\cite{hu2026ideal}.
We therefore combine encoded mask features with geometric support from binary occupancy, retaining spatiotemporal context while localizing mask evidence.

\textbf{Representation and anchoring.} To align mask conditioning with source representations, we encode both through the frozen map $\Phi=\phi\circ\mathcal Q\circ\mathcal E$, including input normalization and spatiotemporal flattening. Since an all-zero mask does not encode to zero, we anchor mask evidence to the encoding of an empty mask of the same shape:
\begin{equation}
\tZ^{\tX}=\Phi(\tX),\qquad \Delta\tZ^{\tM}=\Phi(\tM_s)-\Phi(\tens{0}).
\label{eq:anchor}
\end{equation}
This subtraction ensures zero mask evidence for an absent edit and represents nonempty masks through changes relative to the shared empty-condition baseline.

\textbf{Geometric support.} Following the VAE's temporal grouping and spatial strides, we use average pooling $\operatorname{AvgPool}$ to map binary occupancy onto the latent grid, then apply spatial max pooling $\operatorname{MaxPool}$ and threshold:
\begin{equation}
\Omega(\tM)=\1\!\left[\operatorname{MaxPool}\!\left(\operatorname{AvgPool}(\tM)\right)\ge0.5\right].
\label{eq:support}
\end{equation}
Max pooling uses unit stride and size-preserving padding, providing a kernel-controlled boundary margin without further downsampling. Complementing empty-mask anchoring, the flattened support $\Omega(\tM)\in\{0,1\}^{L\times1}$ localizes projected mask injection without hard-masking source features or output pixels.

\subsection{Evidence-Guided Generation and Compositional Inference}
\label{sec:generation}

\textbf{Evidence fusion.} To condition the backbone on source content and local style, we project their evidence into a shared $d$-dimensional space using $\mP^{\tX}$ and $\mP^{\tM}$. The mask projection is bias-free to preserve the zero anchored response.
These projections are shared across injection blocks and form $\tU=\mP^{\tX}(\tZ^{\tX})+\Omega(\tM)\odot\mP^{\tM}(\Delta\tZ^{\tM})\in\sR^{L\times d}$, combining source context with geometrically localized mask evidence.
To adapt this shared signal to each block, we compute the channel gate $\vg_k=g_{\mathrm{vis},k}(\tU)+g_{\mathrm{text},k}(p)\in\sR^d$ from visual evidence and text, broadcast it across the $L$ spatiotemporal positions, and inject the gated residual:
\begin{equation}
\tR_k=\tU\odot\vg_k,\qquad
\tH_k=\mathcal G_k\!\left(\tH_{k-1}+\Pi_{\mathrm{vis}}(\tR_k)\right).
\label{eq:injection}
\end{equation}
Here $\mathcal G_k$ is the $k$-th DiT block with output $\tH_k$, and $\Pi_{\mathrm{vis}}$ inserts residuals at existing visual positions without adding tokens. Blocks without evidence injection use $\tR_k=0$.

\textbf{Learning and generation.} With clean source and mask evidence as conditions, GRNStyle predicts target bits from a partially randomized state. We retain GRN's full-video bitwise cross-entropy to jointly supervise stylization and preservation. Training mixes style-specific text with generic mask-following instructions, encouraging the model to read spatially encoded styles while retaining text guidance. At inference, random target bits are iteratively refined under the same conditioning and decoded into video.

\begin{table*}[t]
\centering
\caption{\textbf{Comparison on the LociArt Benchmark.}
The fg/bg suffixes denote stylization of the target mask $M$ and its complement $1-M$.
FT denotes local-stylization fine-tuning, and $*$ denotes training-free methods.
$^{\dagger}$ marks direct mask-compositing methods, for which PSNR is omitted because non-edited regions are copied from the source.
$^{\ddagger}$ marks image-based methods.
Best and second-best results are shown in bold and underlined, respectively.
}
\label{tab:local_stylization_main}
\resizebox{\linewidth}{!}{%
\begin{tabular}{cc|ccccccc|cc}
\toprule
Method
& Venue
& SPS-fg$\uparrow$ & SPS-bg$\uparrow$
& CLIPT-fg$\uparrow$ & CLIPT-bg$\uparrow$
& PSNR-fg$\uparrow$ & PSNR-bg$\uparrow$
& TC$\uparrow$ & Style$\uparrow$ & Preserve$\uparrow$ \\
\midrule
LEAST$^{*\dagger\ddagger}$~\citep{singh2024least}
& arXiv '24
& 0.283 & 0.302 & 0.158 & 0.173
& - & -
& 0.953 & 2.20 & 4.50 \\
VACE-1.3B~\citep{jiang2025vace}
& ICCV '25
& 0.156 & 0.191 & 0.114 & 0.132
& \textbf{32.35} & \textbf{29.17} & 0.981 & 0.05 & \textbf{5.00} \\
UniVST$^{*\ddagger}$~\citep{song2025univst}
& TPAMI '25
& 0.191 & 0.286 & 0.127 & 0.165
& 27.69 & 23.92 & 0.969 & 1.83 & 4.40 \\
RegionRoute$^{\ddagger}$~\citep{chen2026regionroute}
& CVPR '26
& 0.317 & 0.491 & 0.154 & 0.183
& \underline{29.38} & 13.17 & 0.955 & 2.41 & 2.75 \\
StyleGallery$^{*\dagger\ddagger}$~\citep{he2026stylegallery}
& CVPR '26
& 0.230 & 0.325 & 0.146 & 0.183
& - & -
& 0.957 & 3.39 & \underline{4.77} \\
CoinVE-Edit~\citep{long2026coinve}
& arXiv '26
& 0.248 & 0.251 & 0.153 & 0.159
& 20.71 & 16.76 & \underline{0.983} & 3.11 & 3.93 \\
Seedance 2.5~\citep{bytedance2026seedance25}
& Commercial
& 0.347 & 0.358 & \textbf{0.173} & \underline{0.175}
& 18.50 & 18.76 & \textbf{0.984} & 3.98 & 2.40 \\
\midrule
VACE-1.3B (our FT)
& -
& \underline{0.386} & \underline{0.676} & 0.155 & \underline{0.185}
& 25.43 & 25.49 & 0.980 & \underline{4.52} & 4.56 \\
\textbf{GRNStyle-2B (ours)}
& -
& \textbf{0.407} & \textbf{0.705}
& \underline{0.160} & \textbf{0.199}
& 25.92 & \underline{26.15}
& \underline{0.983} & \textbf{4.62} & 4.64 \\
\bottomrule
\end{tabular}%
}
\end{table*}

\textbf{Compositional inference.} This mask-following interface also supports multiple regions through the same codebook. For $R$ disjoint regions $\{\tM^r\}_{r=1}^R$ and learned styles $\{s_r\}_{r=1}^R$, we compose the encodings in \Eqref{eq:polar}:
\begin{equation}
\tM_s^{\mathrm{multi}}(q)=\sum_{r=1}^R \tM_s^r(q),\qquad
\tM^{\mathrm{union}}=\sum_{r=1}^R \tM^r.
\label{eq:composition}
\end{equation}
Here $\tM_s^r$ denotes the polar encoding of region $\tM^r$.
Disjointness ensures a valid style code or empty condition at each position, with $\mathcal R(\tM_s^{\mathrm{multi}}(q))=\sum_r\tM^r(q)\va_{s_r}$ by \Eqref{eq:readout}. We jointly encode the composite field and localize its mask evidence using $\Omega(\tM^{\mathrm{union}})$.
Thus, all regions are generated in a single refinement trajectory under a generic mask-following instruction, without adding channels or conditioning tokens. Qualitative examples illustrate these compositions after single-region, single-style training, without further training on region--style combinations.

\section{Experiments}
\label{sec:exp}

\subsection{Experimental Setup}
\label{sec:exp_setup}

\noindent\textbf{Training.}
We train the 2B GRNStyle model on LociArt-100K with 8 $\times$ NVIDIA H200 GPUs, using all 18 styles unless otherwise specified. We fully fine-tune the backbone and conditioning branches. More training details are provided in Appendix~\cref{app:model_reproduction}.

\textbf{Benchmark.}
We introduce the LociArt Benchmark for local video stylization; it contains 80 source videos and 84 object-mask units. Each video--mask pair is evaluated under two complementary stylization protocols: foreground stylization edits the tracked foreground-object mask $\tM_{\mathrm{obj}}$, while background stylization edits its complement $1-\tM_{\mathrm{obj}}$. The benchmark therefore contains 168 cases in total. More details can be found in Appendix~\cref{app:benchmark_details}.

\textbf{Metrics.}
We use \textit{SPS} (Style Prototype Similarity) to measure whether the edited region matches the requested style using a prototype bank in the CSD~\citep{somepalli2024measuring} feature space. Regional CLIP-T (\textit{CLIPT})~\citep{radford2021learning} measures alignment between the edited region and the canonical style phrase, while \textit{PSNR}~\citep{hore2010image} measures fidelity to the source in the non-target region. \textit{TC} (Temporal Consistency)~\citep{zhang2024avid} measures cosine similarity between CLIP embeddings of adjacent frames. Additionally, we use Gemini 3.6 Flash to provide separate \textit{Style} and \textit{Preserve} scores for regional stylization quality and non-target content preservation. Detailed definitions and implementation protocols are provided in \cref{app:style_metrics,app:temporal_preservation_metrics,app:vlm_details}.

\begin{figure*}[t]
    \centering
    \includegraphics[width=\linewidth]{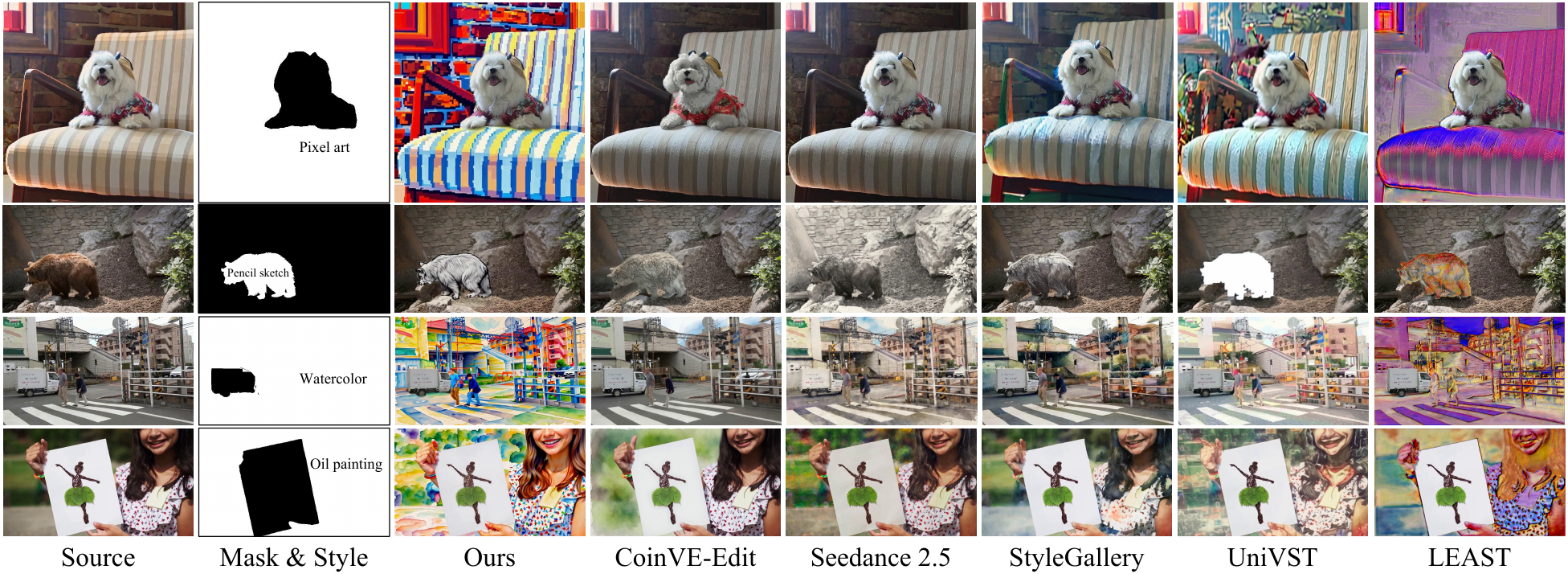}
    \caption{\textbf{Qualitative comparison of local video stylization.}
Examples evaluate stylization inside the editing region, preservation outside it, and temporal consistency.
Foreground and background cases highlight leakage, distortion of preserved content, and boundary inconsistency.
GRNStyle produces localized styles with coherent boundaries and stable non-target content.}
\label{fig:local_stylization_comparison}
\end{figure*}

\subsection{Comparison with Existing Methods}
\label{sec:exp_comparison}

We compare with methods spanning regional image stylization and video editing: LEAST~\citep{singh2024least}, UniVST~\citep{song2025univst}, RegionRoute~\citep{chen2026regionroute}, and StyleGallery~\citep{he2026stylegallery} are image-based methods, whereas VACE-1.3B~\citep{jiang2025vace}, CoinVE-Edit~\citep{long2026coinve}, and the commercial system Seedance 2.5~\citep{bytedance2026seedance25} target video generation or editing. LEAST and StyleGallery directly composite the non-edited region from the source using the mask, while others generate or reconstruct the complete output; accordingly, PSNR is not reported for the former two because source-pixel copying makes it non-comparable. To isolate the contribution of task-specific supervision, we additionally fine-tune VACE-1.3B on LociArt-100K using the same architecture-independent hyperparameters as GRNStyle.

\textbf{Quantitative and perceptual comparison.}
As shown in \cref{tab:local_stylization_main}, GRNStyle achieves the best SPS for both foreground (0.407) and background (0.705) stylization, together with the best background CLIPT (0.199). It also maintains strong preservation, with the second-best background PSNR (26.15 dB) and near-best temporal consistency (TC: 0.983). The VLM scores further indicate a favorable balance between regional stylization and non-target preservation, achieving the highest Style score (4.62) while retaining a high Preserve score (4.64). The highest PSNR values from VACE coincide with its minimal stylization, so they do not indicate a better style--preservation trade-off. In addition, Appendix~\ref{app:global_stylization} shows that GRNStyle remains competitive with SOTA global stylization methods under all-one masks.

\textbf{Qualitative comparison.}
\cref{fig:local_stylization_comparison} presents representative resultsof local video stylization. GRNStyle renders the armchair and room in pixel art while preserving the dog, and adds pencil strokes to the bear without desaturating the surrounding rocks, demonstrating precise regional stylization and faithful preservation. Other methods tend to produce weak edits, spill stylization beyond the target, or lose subject detail. Additional examples are provided in Appendix~\ref{sec:additional_qualitative}.

\subsection{User Study}
\label{sec:exp_user_study}

\begin{figure}[t]
    \centering

    \begin{minipage}[t]{0.44\linewidth}
        \centering
        \includegraphics[width=\linewidth]{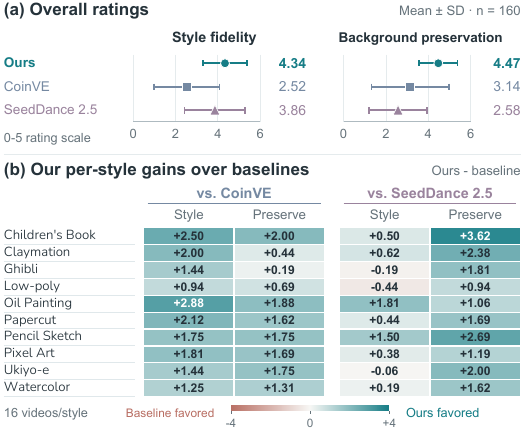}
        \caption{\textbf{Human evaluation results.}
        Values show mean $\pm$ SD; per-style gains are shown for the ten-style subset.}
        \label{fig:human_evaluation}
    \end{minipage}%
    \hfill
    \begin{minipage}[t]{0.54\linewidth}
        \centering
        \includegraphics[width=\linewidth]{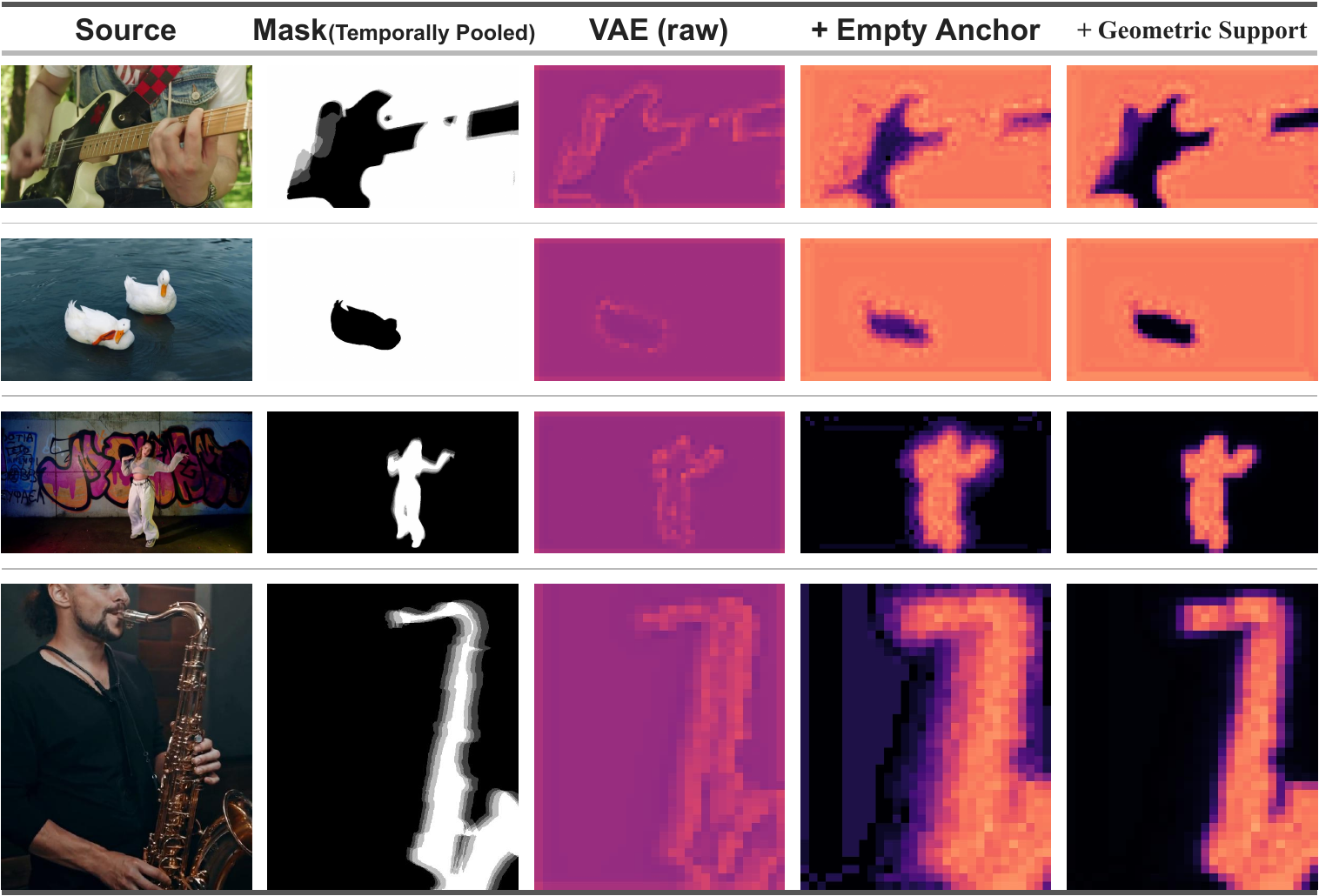}
        \caption{\textbf{Mask-conditioning ablation.}
        Representation--geometry conditioning improves spatial localization over the raw encoding.}
        \label{fig:mask_encoding_ablation}
    \end{minipage}

\end{figure}

We conducted a user study with 20 participants on a 160-case subset (16 cases per style), with each case receiving one rating from this participant pool. Since the study evaluates generated videos, we compare GRNStyle only with the two video-based methods, CoinVE-Edit and Seedance 2.5. As shown in \cref{fig:human_evaluation}(a), GRNStyle achieves the highest mean ratings for style fidelity (4.34) and background preservation (4.47), indicating stronger artistic expression together with better scene preservation. \cref{fig:human_evaluation}(b) further shows consistent benefits across styles: GRNStyle outperforms CoinVE-Edit on both criteria for all ten styles, improves preservation over Seedance 2.5 for every style, and improves style fidelity for seven styles.

\subsection{Ablation of Mask Conditioning}
\label{sec:mask_ablation}

We evaluate two complementary mask-conditioning strategies: RG (representation--geometry conditioning), which combines empty-mask anchoring with raw-mask geometric support, and Polar conditioning, which encodes an explicit regional style identity through a polar representation.

\cref{tab:mask_components} shows that the baseline performs worst overall. RG substantially improves PSNR while leaving SPS and CLIPT nearly unchanged, indicating improved preservation without sacrificing stylization. As illustrated in \cref{fig:mask_encoding_ablation}, empty-mask anchoring removes the reference response to an empty input, while geometric support suppresses spatial spillover. Polar encoding supplies an explicit regional style identity, yielding a clear SPS improvement and a modest PSNR gain. Combining the two components achieves the strongest overall trade-off between style conditioning and spatial control. The appendix analysis in \cref{app:mask_sensitivity} further shows that GRNStyle remains robust to imperfect masks.

\begin{table}[t]
\centering
\caption{\textbf{Component ablation.} RG: representation--geometry conditioning. Polar: polar style-aware mask.}
\label{tab:mask_components}
\begin{tabular}{lccc}
\toprule
Configuration & SPS$\uparrow$ & CLIPT$\uparrow$ & PSNR$\uparrow$ \\
\midrule
Base & 0.541 & 0.180 & 20.09 \\
Base + Polar & 0.553 & 0.181 & 21.34 \\
Base + RG & 0.544 & 0.179 & 25.07 \\
\textbf{GRNStyle} & 0.554 & 0.181 & 25.78 \\
\bottomrule
\end{tabular}
\end{table}

\subsection{Learning New Styles without Replay}
\label{sec:exp_incremental}

To examine whether a predefined polar codebook limits expansion to new styles, we extend a separately trained 10-style model to 15 styles using only the five new classes and no replay of the original data. The polar and binary variants follow the same protocol, and retention is measured against their respective 10-class checkpoints.

\begin{table}[t]
\centering
\caption{\textbf{Replay-free style expansion from 10 to 15 classes.}
Expanded models are trained only on the five new classes.
Retention and acquisition are evaluated on 1,680 original-style and 840 new-style cases, respectively.}
\label{tab:style_expansion}
\small
\resizebox{\linewidth}{!}{%
\begin{tabular}{lcccccc}
\toprule
& \multicolumn{3}{c}{10 classes (old)} & \multicolumn{3}{c}{5 classes (new)} \\
\cmidrule(lr){2-4}\cmidrule(lr){5-7}
Configuration & SPS$\uparrow$ & CLIPT$\uparrow$ & PSNR$\uparrow$ & SPS$\uparrow$ & CLIPT$\uparrow$ & PSNR$\uparrow$ \\
\midrule
Binary, 10 classes & 0.547 & 0.177 & 25.42 & --- & --- & --- \\
Binary, 15 classes & 0.445\olddrop{0.102} & 0.158\olddrop{0.019} & 25.43\oldgain{0.01} & 0.588 & 0.197 & 25.39 \\
\midrule
Polar, 10 classes & 0.553 & 0.180 & 25.75 & --- & --- & --- \\
Polar, 15 classes & 0.531\olddrop{0.022} & 0.172\olddrop{0.008} & 25.77\oldgain{0.02} & 0.590 & 0.199 & 25.74 \\
\bottomrule
\end{tabular}
}
\end{table}

\Cref{tab:style_expansion} shows that polar conditioning supports this 50\% vocabulary expansion with substantially less forgetting: old-class SPS drops by 0.022, compared with 0.102 for binary masks, while the corresponding CLIPT drops are 0.008 and 0.019; PSNR remains stable in both cases. The new-class scores are also slightly higher across all metrics, indicating that improved retention does not hinder adaptation. These results support the extensibility of explicit style identities: new styles can be added while better preserving existing capabilities, without replaying the original data.

\subsection{Effect of the Data Construction Pipeline}
\label{sec:exp_data_pipeline}

\begin{figure*}[t]
    \centering
    \includegraphics[width=\linewidth]{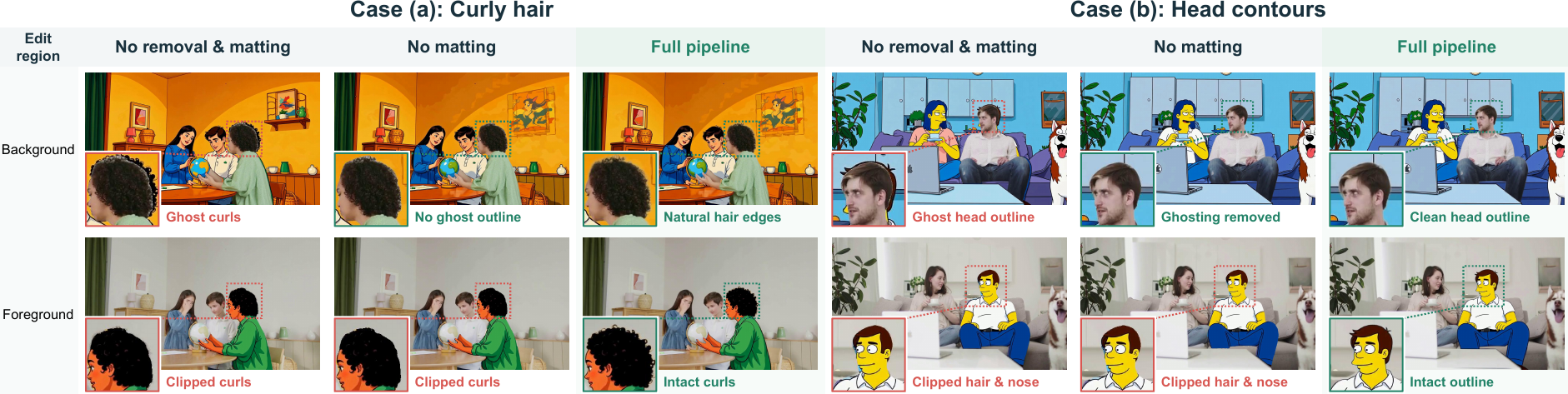}
\caption{\textbf{Data-construction ablation.}
For curly hair (a) and head contours (b), columns compare direct compositing, removal without matting, and the full pipeline.
Rows show background- and foreground-stylized targets; insets highlight residual silhouettes, clipped contours, and boundary quality.}
    \label{fig:data_pipeline_ablation}
\end{figure*}

To assess the contribution of each data-construction step, we compare direct mask compositing, using object removal but no matting, and the complete pipeline for both foreground and background stylization. As shown in \cref{fig:data_pipeline_ablation}, background editing with direct compositing can leave ghost curls or head outlines when stylized subjects extend beyond their source silhouettes; object removal eliminates these artifacts. For foreground editing, source-aligned masks may clip stylized curls, hair tips, and noses, while removal alone cannot resolve the mismatch. Matting follows the actual foreground contour, preserving fine boundaries and providing cleaner supervision for style-induced shape changes.

\section{Conclusion}
\label{sec:conclusion}

We introduced LociArt, a unified framework comprising automated
data construction, the LociArt-100K dataset, GRNStyle, and a
region-aware benchmark. Built on GRN, GRNStyle combines
representation--geometry conditioning with polar style codes to
balance spatial control and temporal context while preserving
non-target content, outperforming the evaluated baselines on the
LociArt Benchmark.
Category-aware polar style-aware masks further enable zero-shot multi-region
compositions from single-region, single-style supervision, reducing
the need for exhaustive paired data.
Limitations and future works are discussed in Appendix~\cref{app:limitations}.

\section*{Acknowledgements}
This work uses the VideoMaMa model (\url{https://github.com/cvlab-kaist/VideoMaMa}), licensed under CC BY-NC 4.0.
The Ditto dataset and model (\url{https://github.com/EzioBy/Ditto}), licensed under CC BY-NC-SA 4.0.
The authors confirm that all uses of the above resources are strictly for academic research purposes and not for any commercial application.

\bibliography{main}

\clearpage
\appendix
\newpage

\section*{Appendix}
The supplementary material is organized as follows:
\begin{itemize}
    \item \textbf{\ref{sec:data_pipeline_details}.}~\hyperref[sec:data_pipeline_details]{Detailed Data Pipeline} (p.~\pageref{sec:data_pipeline_details}): data construction, filtering, verification, and statistics.
    \item \textbf{\ref{app:style_encoding}.}~\hyperref[app:style_encoding]{Style-Code Design and Local Readout Analysis} (p.~\pageref{app:style_encoding}): circular codebook, occupancy normalization, and readout analysis.
    \item \textbf{\ref{app:eval_protocol}.}~\hyperref[app:eval_protocol]{Details of the Evaluation Protocol} (p.~\pageref{app:eval_protocol}): benchmark construction, localized metrics, and VLM evaluation.
    \item \textbf{\ref{app:model_reproduction}.}~\hyperref[app:model_reproduction]{Model Reproduction Details} (p.~\pageref{app:model_reproduction}): training configuration, evidence injection, and checkpoint validation.
    \item \textbf{\ref{app:efficiency}.}~\hyperref[app:efficiency]{Inference Efficiency} (p.~\pageref{app:efficiency}): latency and GPU-memory comparison.
    \item \textbf{\ref{app:global_stylization}.}~\hyperref[app:global_stylization]{Global Stylization Comparison} (p.~\pageref{app:global_stylization}): all-one-mask comparison with global stylization methods.
    \item \textbf{\ref{app:mask_sensitivity}.}~\hyperref[app:mask_sensitivity]{Robustness to Imperfect Masks} (p.~\pageref{app:mask_sensitivity}): five mask perturbations at two severity levels.
    \item \textbf{\ref{app:limitations}.}~\hyperref[app:limitations]{Limitations and Future Work} (p.~\pageref{app:limitations}): current limitations and future directions.
    \item \textbf{\ref{sec:additional_qualitative}.}~\hyperref[sec:additional_qualitative]{Additional Qualitative Results} (p.~\pageref{sec:additional_qualitative}): additional foreground and background examples.
\end{itemize}

\section{Detailed Data Pipeline}
\label{sec:data_pipeline_details}

All videos are standardized to 81 frames at 720p resolution. Frame correspondence is maintained throughout object removal, global stylization, matting, and compositing, so source, removed, stylized, and alpha videos remain aligned. Both Ditto~\citep{bai2026scaling} and our internal fine-tuned VACE~\citep{jiang2025vace} are run on the source and object-removed videos; each stylizer produces candidate outputs for the subsequent quality checks.

\subsection{Object Inventory and Mask Filtering Details}
\label{sec:stage1_details}

This stage identifies candidate foreground objects and obtains reliable masks for object removal and subsequent local composition. We first build an object inventory, track the selected instances, and then filter their masks for semantic correctness and temporal coverage.

\paragraph{Object inventory.}
Qwen3-VL-32B~\citep{bai2025qwen3} analyzes each source video to identify prominent foreground objects, excluding broad background regions such as sky, roads, and walls. Each inventory entry contains a concise category label and an appearance- and location-based description. The category label provides the prompt for segmentation.

\paragraph{Instance handling.}
Using these category prompts, SAM3~\citep{carion2026sam} produces temporally tracked instance masks. When several objects share a category, their masks remain separate if their identities can be reliably tracked, preserving object-specific control for the later removal and compositing steps.

\paragraph{Mask filtering.}
We then use Qwen3-VL-32B~\citep{bai2025qwen3} to reject missing, flickering, or incorrectly assigned masks. Additional checks require nonempty masks in more than 40\% of frames and reject masks whose maximum per-frame area exceeds 85\% of the frame. Temporal coverage counts all nonempty frames, not the length of a continuous track. These production thresholds select masks with sufficient visibility and a bounded foreground extent before object removal.

\subsection{Global Stylization Quality Filtering Details}
\label{sec:stage2_details}

The source and object-removed videos provide complementary content for local composition, so both stylizations must preserve their inputs while expressing the requested style. We assess them through two cascaded gates with a shared sampling and scoring protocol, retaining only candidates that satisfy the acceptance criteria below.

\paragraph{Cascaded evaluation.}
The source-video pair is evaluated first; failure at this gate stops processing for the candidate. Passing candidates proceed to evaluation of the object-removed pair, and both pairs must pass before matting and compositing. To avoid repeated evaluation, source-gate results are reused for candidates sharing the same source, style instruction, and actual stylized output. Removal-gate results are cached separately for each object-removed video, style instruction, and corresponding stylized output. Distinct removal inputs or generated outputs require separate evaluations, even when the source video and style are shared.

\paragraph{Frame sampling and scoring.}
At either gate, we sample frame indices 0, 20, 40, 60, and 80 from the aligned 81-frame videos. GPT-5.5 receives ten images in chronological paired order, with an original and a stylized frame at each time point. It scores style adherence, content preservation, structural quality, style consistency, and temporal quality on a 1--5 scale.

\paragraph{Acceptance criteria.}
The resulting five scores must reach $4$, $4$, $3$, $4$, and $4$, respectively. In addition, a pair is rejected if it introduces salient objects, removes or duplicates objects, changes semantics or identity, inserts text or logos, or exhibits major layout changes, severe deformation, or ghosting. The total score is reported for reference only and imposes no additional passing threshold.

\paragraph{Evaluation prompt.}
The following prompt (\cref{lst:vlm_filter_prompt}) implements the same sampling, scoring, and acceptance protocol for both gates, with the corresponding input video and stylized output supplied in each case.

\begin{figure*}
    \centering
    \includegraphics[width=\linewidth]{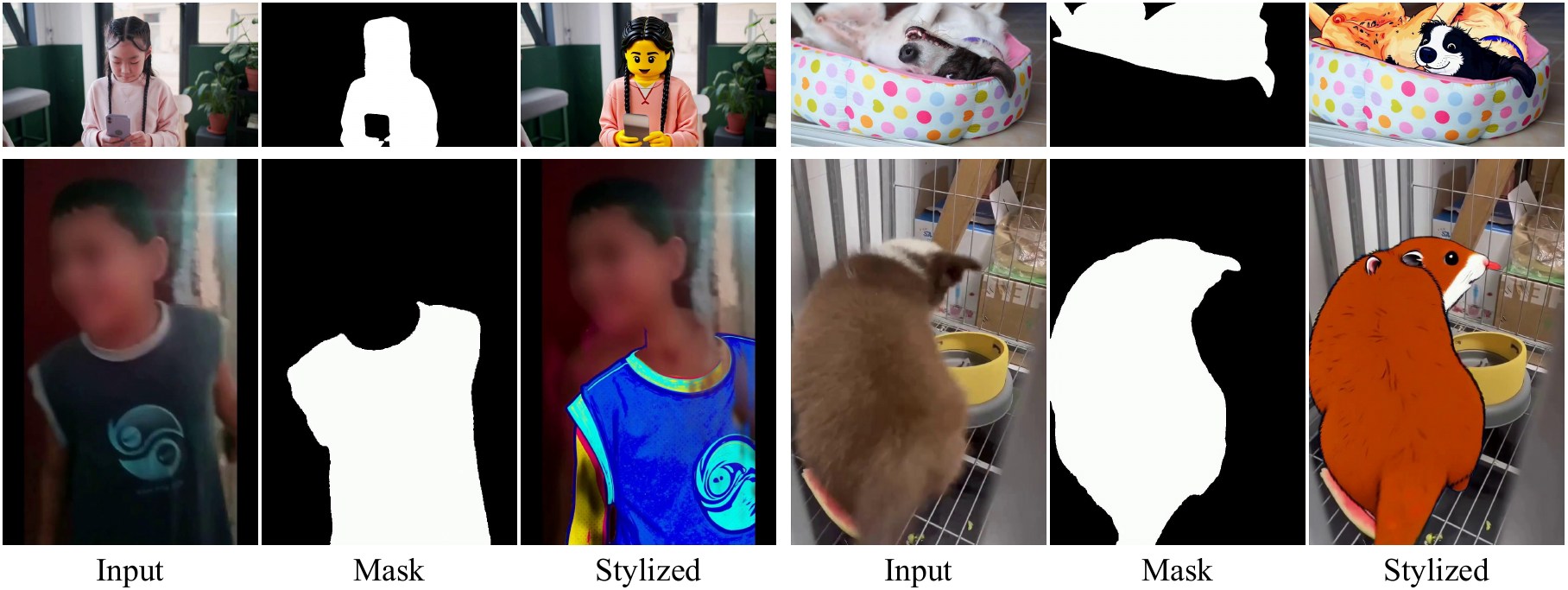}
    \caption{Representative rejection cases from the data pipeline.}
    \label{fig:pipeline_badcase}
\end{figure*}

\subsection{Human Verification Details}
\label{sec:stage4_details}

After matting and compositing, human review checks the completed local edits for failures that global stylization filtering may miss. Reviewers assess each source--mask--instruction--target example against the criteria below, then retain qualified pairs to form the final dataset.

\begin{lstlisting}[
    basicstyle=\ttfamily\scriptsize,
    breaklines=true,
    breakatwhitespace=false,
    breakindent=0pt,
    columns=fullflexible,
    keepspaces=true,
    frame=single,
    caption={VLM prompt for global stylization quality filtering.},
    label={lst:vlm_filter_prompt}
]
You are a strict content-preservation evaluator for a full-video stylization result.
Target production style: {production_prompt}
Canonical style category: {style_category}
The input videos contain 81 aligned frames (indices 0 through 80). You receive 10 images from five uniformly sampled time points in chronological paired order:
- Images 1 and 2: ORIGINAL frame 0 and STYLIZED frame 0
- Images 3 and 4: ORIGINAL frame 20 and STYLIZED frame 20
- Images 5 and 6: ORIGINAL frame 40 and STYLIZED frame 40
- Images 7 and 8: ORIGINAL frame 60 and STYLIZED frame 60
- Images 9 and 10: ORIGINAL frame 80 and STYLIZED frame 80

Each original/stylized pair is the same time point. Compare every pair and use all five time points. Do not decide from the first pair alone.
Evaluate the requested style while preserving semantic content strictly. Content preservation includes the same people, animals, objects, object count, identities, roles, actions, interactions, scene layout, camera view, and major background structures.
Do not penalize normal style transformations such as changed colors, simplified details, stylized facial features that preserve identity, painterly/cartoon/anime/clay/pixel/vector rendering, or moderate texture simplification.
However, be conservative: if an object identity, person identity, object count, action, or scene layout cannot be confidently matched between an original/stylized pair, score content_preservation below 4 and reject. Do not assume an ambiguous or redesigned object is preserved.
Reject a new salient person/object, disappearance or duplication, an object changing semantic category, identity/role/gender change, action transfer, major layout change, inserted text/logo/graphic, severe deformation, or ghosting.
Judge style adherence against the explicitly requested target style, not against generic stylization or photorealism. Examples:
- Low Poly requires visible facets/polygonal shading.
- Pixel Art requires block-like pixels and a limited palette.
- Claymation requires plasticine material and rounded clay forms.
- Oil Painting requires painterly brushwork.
- Watercolor requires translucent washes.
- Anime or Ghibli requires recognizable hand-drawn animation treatment.
- Other styles must show their characteristic visual language.

Score exactly these five dimensions from 1 to 5:
- style_adherence
- content_preservation
- structural_quality
- style_consistency
- temporal_quality
Passing criteria:
- style_adherence >= 4
- content_preservation >= 4
- structural_quality >= 3
- style_consistency >= 4
- temporal_quality >= 4
- hard_reject_reasons must be empty

Report total as the sum of the five scores for reference only; it is not an additional passing criterion.
Return ONLY one valid JSON object. Do not output analysis, Markdown, or any text outside the JSON object:
{
  "pass": false,
  "style_adherence": 1,
  "content_preservation": 1,
  "structural_quality": 1,
  "style_consistency": 1,
  "temporal_quality": 1,
  "total": 5,
  "hard_reject_reasons": [],
  "reason": "Brief evidence-based explanation"
}
\end{lstlisting}

\paragraph{Verification requirements.}
A qualified sample must satisfy four requirements: (i) the requested style is expressed while semantic content, object identities and counts, actions, and scene layout remain consistent with the source; (ii) non-target content is preserved without unintended style spillover; (iii) masks and compositing boundaries remain coherent, without incomplete regions or visible unstylized remnants; and (iv) the video is temporally consistent, without frame breakage or flicker. These checks distinguish unintended content changes from valid artistic transformations: style-consistent changes in color, texture, detail, and object contours are permitted.

\paragraph{Acceptance and archiving.}
Candidates meeting all four requirements are labeled qualified and retained; those failing the checks are rejected and discarded. Typical rejection cases include weak or incorrect style, semantic or identity changes, inserted or missing objects, unintended changes to preserved content, incomplete masks, and temporal artifacts. Representative examples rejected during human verification are shown in~\cref{fig:pipeline_badcase} and our demo page: \url{https://higherhu.github.io/lociart/}.

\begin{figure*}[t]
    \centering
    \begin{subfigure}[b]{0.38\linewidth}
        \centering
        \includegraphics[width=\linewidth]{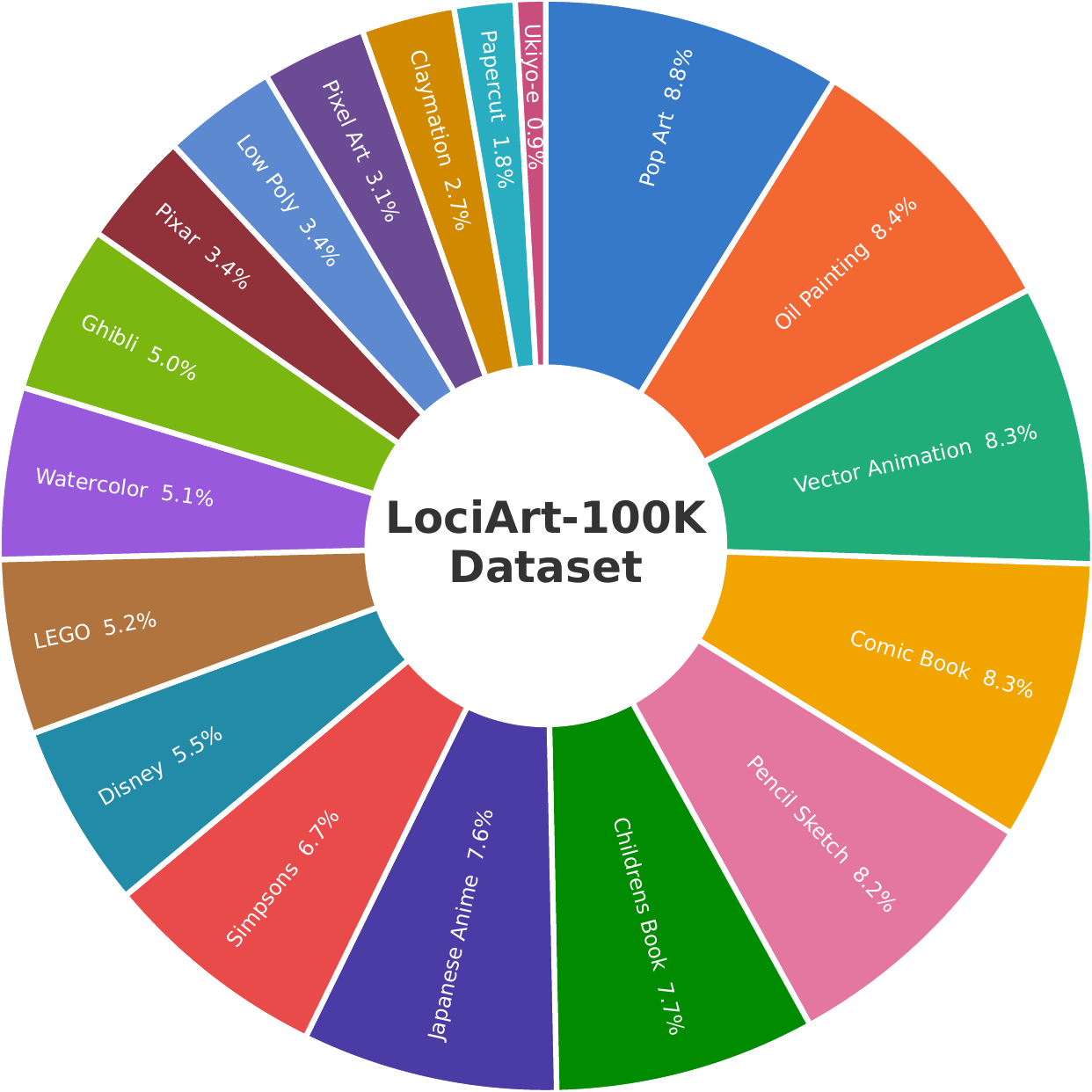}
        \caption{Style-category distribution}
        \label{fig:data_style_class}
    \end{subfigure}
    \hfill
    \begin{subfigure}[b]{0.61\linewidth}
        \centering
        \includegraphics[width=\linewidth]{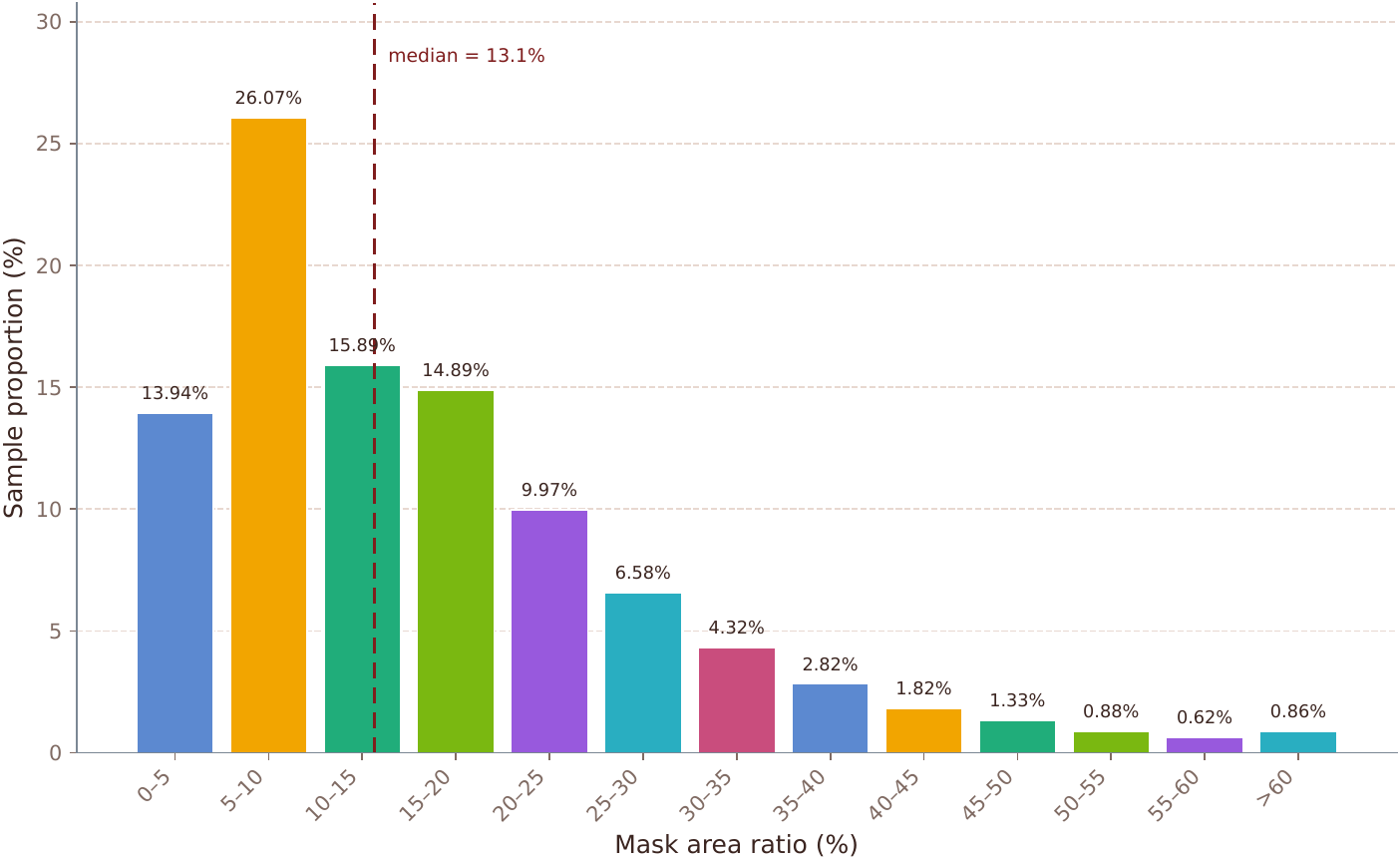}
        \caption{Foreground-object mask area ratios}
        \label{fig:data_mask_area}
    \end{subfigure}
    \caption{\textbf{Statistics of the LociArt-100K Dataset.} The 100K qualified pairs cover 18 style categories. Area ratios refer to foreground object masks $\tM_{\mathrm{obj}}$, not the complementary background editing masks.}
    \label{fig:data_statistics}
\end{figure*}

\paragraph{Production statistics.}
The pass rate is 52.4\% for global stylization VLM filtering and 65\% for human verification of the resulting local composites. These rates are measured at their respective filtering stages. The final dataset contains 100K qualified video pairs covering 18 style categories. \Cref{fig:data_statistics} summarizes the style distribution and foreground object mask areas. The area histogram characterizes $\tM_{\mathrm{obj}}$, not its complementary background editing mask, and therefore does not represent the distribution of all editing regions.

\section{Style-Code Design and Local Readout Analysis}
\label{app:style_encoding}

\subsection{Codebook Assignment and Input Normalization}
\label{app:codebook_normalization}

We use a fixed 100-slot circular codebook. For $K\le100$, let $\nu:\{1,\ldots,K\}\rightarrow\{0,\ldots,99\}$ assign each style to a distinct slot, with $\theta_s=2\pi\nu(s)/100$. Well-spaced slots are assigned deterministically, and the mapping remains fixed across training and inference. The resulting directions satisfy $\|\va_s\|_2=1$, distinguishing styles by angle rather than magnitude, analogous to the equal-norm one-hot bit codes in GRN~\citep{han2026grn}.

Following the main-text definition in \Eqref{eq:polar}, the style-aware mask is
\begin{equation}
\tM_s(q)=\tM(q)\left(\frac{1+a_{s,1}}{2},\frac{1+a_{s,2}}{2},1\right)
\label{eq:app_polar_encoding}
\end{equation}
The VAE input normalization $x\mapsto2x-1$ therefore yields
$c_s=(a_{s,1},a_{s,2},1)$ at edited positions and
$e=(-1,-1,-1)$ at unedited positions. Thus, the normalized third
channel retains the same binary occupancy cue for every style, while
the first two channels encode its direction.

\subsection{Why Circular Rather Than Spherical Codes?}
\label{app:circle_vs_sphere}

A three-coordinate spherical code offers greater class-packing capacity, but uses every channel for identity. Local stylization also needs a stable distinction between edit and keep. We therefore reserve the third channel for occupancy and use the first two for style directions. The shared occupancy channel avoids requiring edit/keep status to be inferred from class-dependent coordinates.

\paragraph{A larger empty-condition margin.}
Reserving this occupancy axis keeps style codes away from the empty condition. For the circular and origin-centered unit-sphere alternatives,
\begin{equation}
\begin{aligned}
\min_{\theta\in[0,2\pi)}
\|(\cos\theta,\sin\theta,1)-e\|_2
&=\sqrt{7-2\sqrt{2}}\approx2.04,\\
\min_{v\in\mathbb R^3:\,\|v\|_2=1}\|v-e\|_2
&=\sqrt{3}-1\approx0.73.
\end{aligned}
\label{eq:app_empty_margin}
\end{equation}
The circular representation therefore keeps every style farther from the empty condition than the spherical alternative. For the 100-slot codebook, the smallest distance between adjacent normalized style directions is $2\sin(\pi/100)\approx0.063$; the empty-condition margin is the larger value $\sqrt{7-2\sqrt{2}}\approx2.04$.

\paragraph{A shared cue for the frozen encoder.}
These coordinates denote category identity, not visual color. The pretrained VAE nevertheless processes them through learned RGB features, so greater spherical spacing alone does not resolve the encoding problem. A binary occupancy channel supplies a common boundary cue across styles, while class-dependent variation remains confined to the other two channels. Our 100-slot VAE/HBQ validation found no exact token collisions. We therefore favor explicit occupancy and empty-condition separation over additional spherical capacity.

\subsection{Interaction Error and Exact Local Readout}
\label{app:readout}

This section isolates the style--region interaction in~\Eqref{eq:readout}. We compare an additive readout of separate conditions with a linear readout of the polar field, using the same region and style information.

\paragraph{Setup.}
At one pixel position, write $m=\tM(q)\in\{0,1\}$ and let $\va_s\in\mathbb R^2$ be the unit style direction in~\Eqref{eq:polar}. The desired local instruction is $m\va_s$. To separate representation from dataset-specific correlations, consider independent $m\sim\operatorname{Bernoulli}(\beta)$ and $s\sim\pi$, where $0<\beta<1$ and $\pi_s$ is the prior probability of style $s$, and define
$\vmu=\sum_s\pi_s\va_s$. For additive readouts $f(m,s)=f_M(m)+f_S(s)$, the component functions may be arbitrary. We measure squared instruction error $\mathcal J(f)=\mathbb E\|m\va_s-f(m,s)\|_2^2$.

\paragraph{Optimal additive error.}
The best additive readout and its error are
\begin{align}
f^\star(m,s)
&=m\vmu+\beta(\va_s-\vmu),
\label{eq:app_additive_optimum}\\
\min_{f=f_M+f_S}\mathcal J(f)
&=\beta(1-\beta)\bigl(1-\|\vmu\|_2^2\bigr).
\label{eq:app_additive_error}
\end{align}
To see this, decompose $m\va_s=f^\star(m,s)+(m-\beta)(\va_s-\vmu)$. The residual has zero conditional mean given either $m$ or $s$, so it is orthogonal to every additive readout. Consequently, $\mathcal J(f)=\mathcal J(f^\star)+\mathbb E\|f-f^\star\|_2^2$.
Independence and $\|\va_s\|_2=1$ then give~\Eqref{eq:app_additive_error}. The factor $1-\|\vmu\|_2^2$ is the variance of the style directions. The minimum is positive when at least two distinct styles have positive probability. It is at most $\beta(1-\beta)\le1/4$, with equality at $1/4$ when $\beta=1/2$ and $\vmu=0$.

\paragraph{Exact polar readout.}
For the same inputs, the encoded pixel $\tM_s(q)$ defined in \Eqref{eq:app_polar_encoding} satisfies
\begin{equation}
\begin{pmatrix}
2&0&-1\\
0&2&-1
\end{pmatrix}
\tM_s(q)=m\va_s,
\quad
\mathbb E\|\mathcal R(\tM_s(q))-m\va_s\|_2^2=0.
\label{eq:app_polar_exact}
\end{equation}
Polar encoding therefore supplies the interaction explicitly, without adding a new condition. This is a distinction between readout classes, not an error bound on a nonlinear generator: a nonlinear network can learn the interaction from separate inputs, whereas our representation computes it before encoding.

\section{Details of the Evaluation Protocol}
\label{app:eval_protocol}

\begin{figure*}[t]
    \centering
    \begin{subfigure}[b]{0.50\linewidth}
        \centering
        \includegraphics[width=\linewidth]{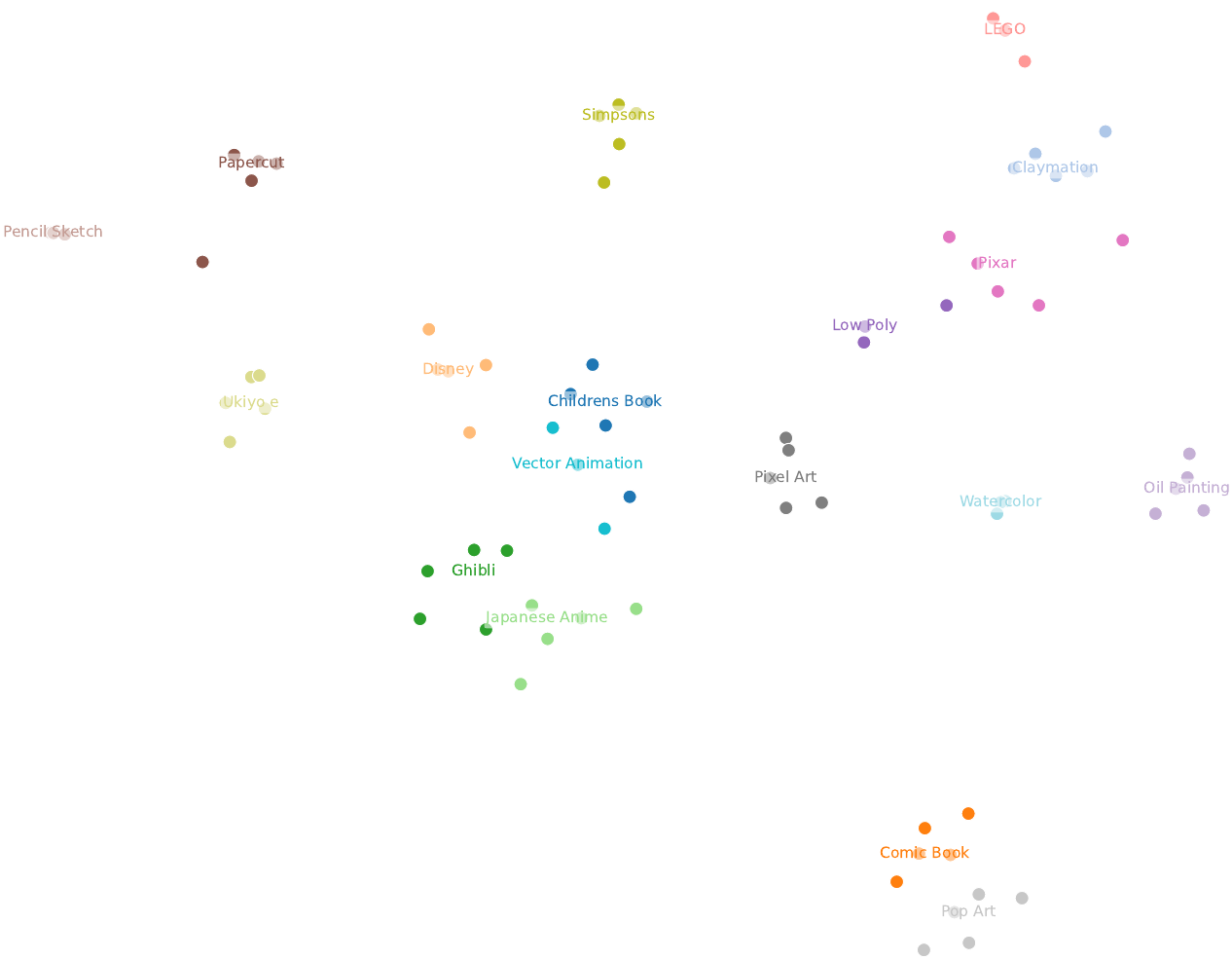}
        \caption{Style prototype visualization.}
        \label{fig:sps_prototype_tsne}
    \end{subfigure}
    \hfill
    \begin{subfigure}[b]{0.49\linewidth}
        \centering
        \includegraphics[width=\linewidth]{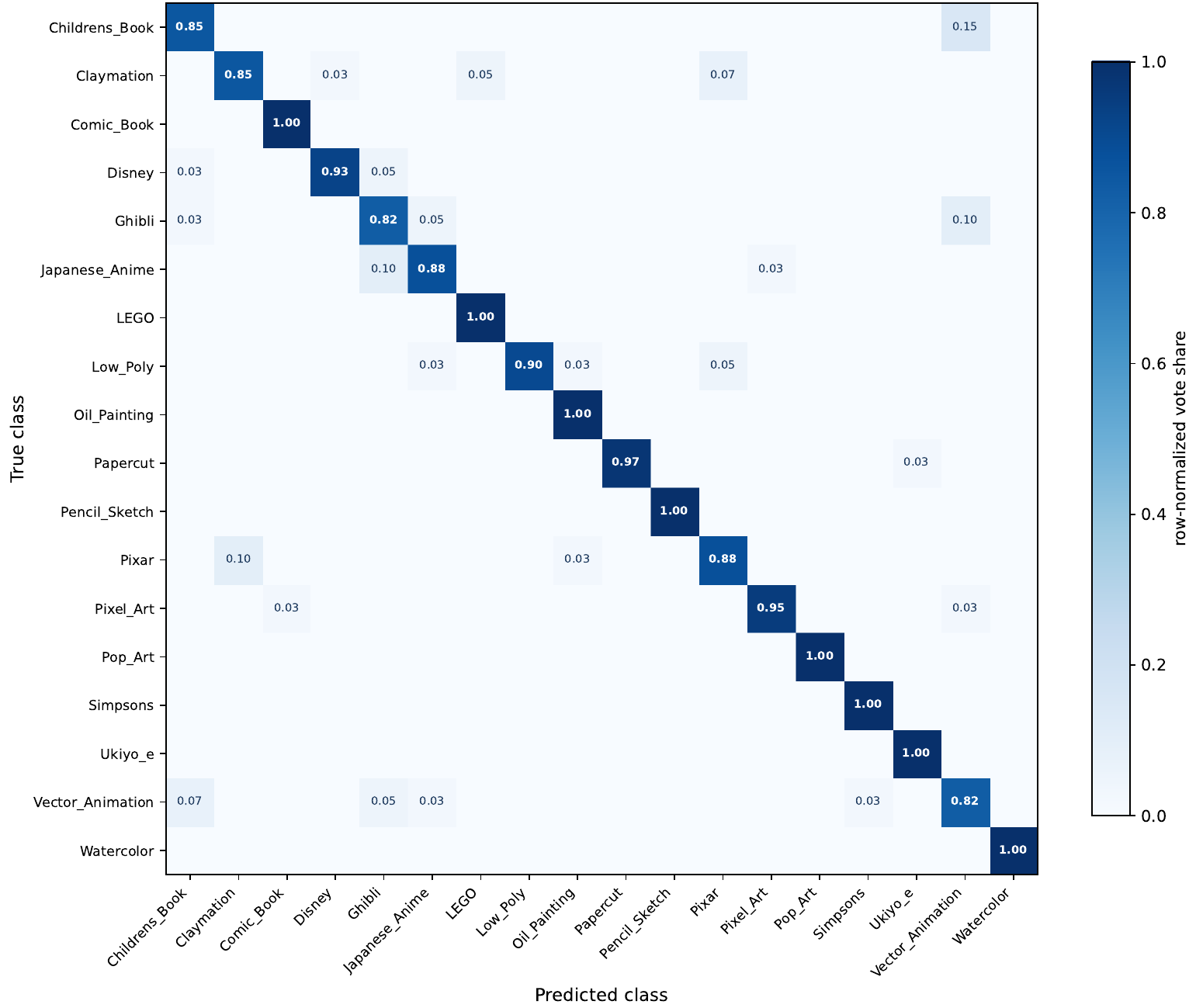}
        \caption{Nearest-prototype classification.}
        \label{fig:sps_confusion}
    \end{subfigure}
    \caption{\textbf{SPS prototype visualization and validation.}
(a) t-SNE visualization of 80 CSD prototypes across 18 styles.
(b) Row-normalized confusion matrix under leave-one-source-out validation, achieving 93.61\% macro-averaged accuracy.}
    \label{fig:sps_validation}
\end{figure*}

We first describe the construction and evaluation units of the LociArt Benchmark, then assess local stylization along three complementary axes: regional style, measured by Style Prototype Similarity (SPS) and regional CLIPT; temporal consistency and non-target preservation, measured by temporal consistency (TC) and regional PSNR; and full-video perceptual quality, measured by the VLM Style and Preserve scores.

\subsection{Benchmark Construction and Evaluation Units}
\label{app:benchmark_details}

The benchmark contains 80 source videos and 84 object-mask units, each with 81 aligned frames. Videos with two selected subjects contribute separate object-mask units. Of the 80 videos, 13 come from the rejected Ditto~\citep{bai2026scaling} pool and were excluded from our training data, 46 come from the DAVIS 2017 training split~\citep{pont20172017}, and 21 come from Pexels\footnote{\url{https://www.pexels.com/}}. The foreground area ratios range from 0.02 to 0.58, and 29 units contain human subjects. We denote the tracked foreground-object mask by $\tM_{\mathrm{obj}}$ and the directional editing mask by $\tM$, with $\tM=\tM_{\mathrm{obj}}$ for foreground editing and $\tM=1-\tM_{\mathrm{obj}}$ for background editing. Throughout the metric definitions below, $N$ denotes the number of frames in the evaluated aligned sequence and $n\in\{1,\ldots,N\}$ indexes a frame; for benchmark clips, the nominal sequence length is 81.

\paragraph{Styles and comparison coverage.}
In the quantitative and perceptual comparison, all methods receive the same input videos for the same target style and follow the same frame-alignment protocol for shared cases. Table~\ref{tab:eval_style_vocabulary} lists the 18-style vocabulary and the ten-style comparison subset (\texttt{bench10}). Each object-mask unit supports foreground and background editing, so the complete ten-style grid contains $84\times10\times2=1{,}680$ results (168 cases, 10 styles).

\begin{table}[t]
\centering
\caption{\textbf{Canonical style phrases for regional CLIPT.} A check mark identifies the ten-style comparison subset.}
\label{tab:eval_style_vocabulary}
\small
\begin{tabular}{llc}
\toprule
Style category & Canonical phrase & in \texttt{bench10} \\
\midrule
Children's Book & children's book illustration & $\checkmark$ \\
Claymation & claymation & $\checkmark$ \\
Comic Book & comic book & \\
Disney & Disney animation & \\
Ghibli & Studio Ghibli anime & $\checkmark$ \\
Japanese Anime & Japanese anime & \\
LEGO & LEGO brick & \\
Low Poly & low-poly 3D & $\checkmark$ \\
Oil Painting & oil painting & $\checkmark$ \\
Papercut & papercut & $\checkmark$ \\
Pencil Sketch & pencil sketch & $\checkmark$ \\
Pixar & Pixar 3D animation & \\
Pixel Art & pixel art & $\checkmark$ \\
Pop Art & pop art & \\
Simpsons & The Simpsons cartoon & \\
Ukiyo-e & Ukiyo-e woodblock print & $\checkmark$ \\
Vector Animation & flat vector animation & \\
Watercolor & watercolor painting & $\checkmark$ \\
\bottomrule
\end{tabular}%
\end{table}

\subsection{Regional Style Metrics}
\label{app:style_metrics}

\subsubsection{Localized Region Construction}
\label{app:localized_region}
Both SPS and regional CLIPT operate on a localized region rather than the full output frame. Let $\tY$ be an output video and $\tM_{\mathrm{obj}}$ its tracked foreground-object mask. VideoMaMa~\citep{lim2026videomama} mask-guided matting of the output yields a foreground alpha video $\tA_{\mathrm{out}}$, allowing localization to follow stylization-induced contour changes. \textbf{For foreground stylization}, each frame is cropped to the alpha-support \textit{bounding box with a 2\% margin}. \textbf{For background stylization}, the full frame is retained while \textit{foreground-object tokens are excluded from the region representation}, avoiding an artificial filled foreground area. Thus, SPS and regional CLIPT share the same localization rule and differ only in how the localized region is scored. We denote the resulting CSD region descriptor by $\psi_{\omega}(\tY_n,\tA_{\mathrm{out},n})$ and the corresponding CLIP image embedding by $\chi_{\omega}(\tY_n,\tA_{\mathrm{out},n})$, where $\omega\in\{\mathrm{fg},\mathrm{bg}\}$ specifies the editing direction.

\subsubsection{Style Prototype Similarity (SPS)}
\label{app:sps_details}

SPS evaluates whether an edited region resembles characteristic visual examples of the requested style. Unlike text--image similarity, it compares outputs with a class-specific bank of visual prototypes. SPS is the prototype-based scoring scheme built on the Contrastive Style Descriptors (CSD)~\citep{somepalli2024measuring}.

\paragraph{Prototype construction.}
We collect globally stylized source and object-removed videos that pass the data pipeline's quality gates. Every frame is encoded into a 768-dimensional CSD descriptor. Within each style, we apply $k$-means clustering~\citep{lloyd1982least} to the CSD descriptors. The candidate set $C_s\in\{3,5\}$ is fixed in advance and shared across styles; for each candidate, we compute the mean silhouette score~\citep{rousseeuw1987silhouettes} over the descriptors and retain the partition with the larger score. This data-driven criterion adapts the prototype bank to within-style variation while keeping model selection deliberately small. The resulting bank contains 80 centroids across 18 styles. Multiple prototypes represent within-style variation rather than forcing distinct rendering modes into a single center. \cref{fig:sps_prototype_tsne} visualizes the prototype bank with t-SNE~\citep{van2008visualizing}.

\paragraph{Frame and video scores.}
For target style $s$ and its prototype set $\mathcal P_s$, we compute
\begin{equation}
\begin{aligned}
\rho_{n,s} &= \max_{\vv\in\mathcal P_s}
\operatorname{cos}\!\left(\psi_{\omega}(\tY_n,\tA_{\mathrm{out},n}),\vv\right),\\
\mathrm{SPS} &= \frac{1}{N}\sum_{n=1}^{N}\rho_{n,s}.
\end{aligned}
\label{eq:eval_sps}
\end{equation}
Frame scores are first averaged within each evaluated video; dataset scores then average the corresponding video-level scores, with foreground and background editing reported separately.

\paragraph{Discriminative validation.}
To test whether SPS distinguishes the 18 style categories beyond source-specific appearance, we perform leave-one-source-out validation: all samples derived from the held-out source video are excluded when constructing its reference prototypes. Each frame is assigned the style of its nearest prototype by cosine similarity, and frame predictions are aggregated by video-level majority vote. This yields a macro-averaged classification accuracy of 93.61\%. \cref{fig:sps_confusion} shows the class-wise confusion matrix, complementing the prototype visualization in \cref{fig:sps_prototype_tsne}. Together, these analyses characterize the prototype organization and support SPS's ability to distinguish the evaluated style categories.

\subsubsection{Regional CLIPT}
\label{app:clipt_details}

Regional CLIPT measures alignment with the canonical style phrase, complementing SPS's visual prototypes. Using the same localized region construction as SPS, it extracts image and text features with CLIP ViT-L/14~\citep{radford2021learning}. Let $\chi_{\omega}(\tY_n,\tA_{\mathrm{out},n})$ denote the region image embedding and let $\eta_s$ denote the CLIP text embedding of the canonical phrase for style $s$ in Table~\ref{tab:eval_style_vocabulary}. Then
\begin{equation}
\begin{aligned}
\mathrm{CLIPT}_{\mathrm{reg},n}
&=\operatorname{cos}\!\left(\chi_{\omega}(\tY_n,\tA_{\mathrm{out},n}),\eta_s\right),\\
\mathrm{CLIPT}_{\mathrm{reg}}
&=\frac{1}{N}\sum_{n=1}^{N}\mathrm{CLIPT}_{\mathrm{reg},n}.
\end{aligned}
\label{eq:eval_regional_clipt}
\end{equation}
As with SPS, video-level means are averaged before dataset aggregation. The alpha video defines region selection; it is not itself the RGB image passed to the image encoder. SPS and regional CLIPT therefore differ in their scoring encoders and similarity targets, not in their spatial localization rule.

\subsection{Temporal and Preservation Metrics}
\label{app:temporal_preservation_metrics}

\subsubsection{Temporal Consistency}
\label{app:tc_details}

Following AVID~\citep{zhang2024avid}, TC (Temporal Consistency) uses full-frame image embeddings $\vz_n$ from a CLIP ViT-H/14 tower, a separate and larger encoder than the ViT-L/14 used for regional CLIPT:
\begin{equation}
\mathrm{TC}=\frac{1}{N-1}\sum_{n=1}^{N-1}\operatorname{cos}(\vz_n,\vz_{n+1}),
\quad \vz_n=\operatorname{CLIP}_{\mathrm{img}}(\tY_n).
\label{eq:eval_tc}
\end{equation}
TC summarizes adjacent-frame feature consistency over the aligned output sequence. It is a global measure and does not replace regional style or preservation scores.

\subsubsection{Non-Target Preservation}
\label{app:preservation_details}

Preservation is measured with regional PSNR~\citep{hore2010image} outside the intended edit. For foreground stylization, PSNR is computed on the preserved background region $1-\tM$; for background stylization, it is computed on the preserved foreground region $\tM$. In both cases, the valid region is derived from the foreground alpha in $\tA_{\mathrm{out}}$ on an 8-bit scale: the preserved background core comprises pixels with alpha at most 5, whereas the preserved foreground core comprises pixels with alpha above 200. The anti-aliased transition band in between is excluded.

We compare aligned output and source frames after resizing the source to the output resolution and use the common valid temporal range. For each frame, the RGB reconstruction error is averaged over the valid non-edit pixels and channels, and the resulting errors are aggregated over valid frames before conversion to PSNR. Frames with insufficient valid pixels are skipped; when an alpha estimate is empty, the original mask is used as a fallback.

\subsection{VLM Evaluation}
\label{app:vlm_details}

We use Gemini 3.6 Flash to complement the objective metrics with full-video VLM assessment of two separate quantities: \emph{Style} (\texttt{edit\_region\_style}) and \emph{Preserve} (\texttt{non\_edit\_preservation}). Each is an integer from 0 to 5. The scores are not merged, so strong stylization cannot compensate for damage to non-target content, or vice versa. This evaluation is distinct from the GPT-5.5 data-construction filter, which uses five dimensions on a 1--5 scale.

\paragraph{Inputs and execution.}
The judge receives the source video first and the edited result second, aligned frame by frame and tagged \texttt{SRC} and \texttt{RESULT}. A red contour marks the foreground object in the source. For foreground editing, its interior is the target; for background editing, the exterior is the target. The text instruction explicitly identifies the direction, and the contour and display tags are not scored. We use the English system prompt and rubric below with temperature zero. Responses are parsed as JSON; if parsing fails, the same request is retried once.

\paragraph{Scoring anchors.}
\emph{Style} jointly assesses style correctness, coverage of the edit region, and cross-frame stability. A score of 5 requires all three, 3 indicates visible but imperfect style, incomplete coverage, or occasional flicker, and 0 indicates absent or incorrect style or severe instability. \emph{Preserve} assesses source fidelity and temporal stability in the non-edit region: 5 indicates a close, stable match, 3 allows minor changes or occasional jitter, and 0 indicates substantial repainting or severe drift. The two video-level scores are aggregated separately across cases.

\paragraph{Prompt templates.}
The following template (\cref{lst:eval_vlm_user}) retains the scoring rubric while making the edit direction explicit. For foreground editing, \texttt{\{target\_region\}} is ``the foreground object inside the red outline''; for background editing, it is ``the background outside the red outline.'' The \texttt{\{region\_description\}} field names both the edit and non-edit regions accordingly, and \texttt{\{style\}} is replaced by the requested style.

\begin{lstlisting}[
    basicstyle=\ttfamily\scriptsize,
    breaklines=true,
    breakatwhitespace=false,
    breakindent=0pt,
    columns=fullflexible,
    keepspaces=true,
    frame=single,
    caption={VLM-as-a-judge prompt in our evaluation.},
    label={lst:eval_vlm_user}
]
You are a professional video local-editing judge. This review contains two videos: the first is the source video (the foreground object is outlined in red, with an "SRC" tag in the top-left corner), and the second is the edited result video (tagged "RESULT" in the top-left corner). The red outline and any display differences caused by it as a marker are not scored; judge only the actual visual changes inside the outlined region and in the rest of the frame. Watch both videos in full and compare them segment by segment before scoring. Score strictly based only on what is visible; do not deduct for video encoding noise, and do not let the order of the videos bias you. Output only one line of valid JSON, with no code fences or any other text.
[Task] Local video stylization review
- Edit request: convert only {target_region} in the source video to the {style} style
- Region: {region_description}
[Scoring dimensions] Each dimension is an integer from 0 to 5; score ONLY the following two dimensions:
1. edit_region_style -- quality of the stylization in the edit region (combining style correctness, coverage completeness, and cross-frame stability):
   (i) Stylization: does it present a pure, typical {style} style
   (ii) Coverage: does the style evenly cover the entire edit region (no gaps, no missed spots)
   (iii) Temporal stability: the appearance of the edit region stays stable and coherent over time, with no flicker and no frame-by-frame style drift
   5 = a pure, typical {style} style evenly covering the entire edit region, with no noticeable abnormal flicker or style drift throughout; if any aspect clearly fails, a 5 must not be given
   3 = stylization is present but not faithful to the style, or coverage is incomplete / parts of the region are missed, or occasional flicker
   0 = almost no stylization, the style is completely wrong, or the edit region flickers or drifts severely
2. non_edit_preservation -- preservation of the non-edit region (combining):
   (i) Consistency with the source: at every corresponding moment the non-edit region keeps the content, position, texture, and colors of the source video as closely as possible
   (ii) Temporal consistency: the non-edit region stays stable over time, with no frame-by-frame jitter and no color-shift drift
   5 = the non-edit region matches the source video closely and is temporally stable throughout
   3 = only minor changes, slight color shifts, or local differences, and most of the non-edit region still matches the corresponding source frames; or only occasional jitter
   0 = the content, texture, colors, or details of the non-edit region are visibly repainted or stylized, or severe frame-by-frame jitter appears

[Output] Output only the following JSON and nothing else:
{"edit_region_style":n,"non_edit_preservation":n,"comment":"summarize the most prominent issue in at most 20 English words; write none if there is no obvious issue"}
\end{lstlisting}

\section{Model Reproduction Details}
\label{app:model_reproduction}

This appendix specifies the training configuration of the reported
GRNStyle-2B model, defines the checkpoint validation protocol, and discusses
the observed validation trends.

\subsection{Training Setup}
\label{app:training_setup}

Table~\ref{tab:grnstyle_hyperparameters} summarizes the main training
configuration, which uses 100K paired examples across 18 style categories.
The model is initialized from a pre-trained GRNEdit~\citep{xie2026grnedit}
general-purpose video editor and fully fine-tunes the GRN-2B backbone and
conditioning branches, while the video tokenizer and T5 encoder remain
frozen. This initialization provides a strong global stylization prior, so
optimization primarily adapts the model to mask-guided local control rather
than learning stylization from scratch.

Training applies the all-token bit cross-entropy over the full video. With
probability 0.50, the style-specific instruction is replaced by a generic
mask-following instruction, encouraging the model to read style identity
from the polar style-aware mask while retaining text guidance. The training
loader consumes 61-frame clips at 20 fps.

\begin{table*}[!htpb]
\centering
\caption{Main training configuration for GRNStyle-2B.}
\label{tab:grnstyle_hyperparameters}
\begingroup
\definecolor{grnTableTint}{HTML}{EDF4F5}
\definecolor{grnTableInk}{HTML}{244C56}
\small
\setlength{\tabcolsep}{7pt}
\renewcommand{\arraystretch}{1.10}
\begin{tabularx}{\linewidth}{@{}>{\raggedright\arraybackslash}p{0.48\linewidth} >{\raggedright\arraybackslash}X@{}}
\toprule
\textbf{Hyperparameter} & \textbf{Setting} \\
\midrule
\rowcolor{grnTableTint}[0pt][0pt]\multicolumn{2}{@{}l@{}}{\strut\color{grnTableInk}\textbf{Model and representations}} \\
Backbone & GRN-2B \\
Initialization & Pre-trained GRNEdit \\
Trainable modules & GRN-2B backbone, conditioning branches \\
Transformer blocks / hidden width / heads & 28 / 2,304 / 18 \\
\addlinespace[2pt]
\rowcolor{grnTableTint}[0pt][0pt]\multicolumn{2}{@{}l@{}}{\strut\color{grnTableInk}\textbf{Data and conditioning}} \\
Training pairs / style categories & 100,000 / 18 \\
Training clip length / sampling rate & 61 / 20 fps \\
Aspect-ratio buckets & 15 buckets, $H/W\in[0.333,3.0]$ \\
Class-aware polar mask & Enabled \\
Generic-prompt replacement probability & 0.50 \\
\addlinespace[2pt]
\rowcolor{grnTableTint}[0pt][0pt]\multicolumn{2}{@{}l@{}}{\strut\color{grnTableInk}\textbf{Optimization}} \\
Optimizer / momentum coefficients & AdamW / $(0.9,0.999)$ \\
Weight decay & 0.01 \\
Gradient clipping (global norm) & 0.1 \\
Training objective & All-token bit cross-entropy \\
Mixed precision / TF32 & BF16 / enabled \\
Training duration & 10 epochs \\
\addlinespace[2pt]
\rowcolor{grnTableTint}[0pt][0pt]\multicolumn{2}{@{}l@{}}{\strut\color{grnTableInk}\textbf{Learning-rate schedule}} \\
Initial / peak / minimum learning rate & $10^{-5}$ / $4\!\times\!10^{-5}$ / $2\!\times\!10^{-5}$ \\
Linear warmup & 100 optimizer updates \\
\addlinespace[2pt]
\rowcolor{grnTableTint}[0pt][0pt]\multicolumn{2}{@{}l@{}}{\strut\color{grnTableInk}\textbf{Distributed execution}} \\
Distributed ranks / sharding & 8 / FSDP full sharding \\
Random seed & 1,789,151,675 \\
\bottomrule
\end{tabularx}
\endgroup
\end{table*}

\paragraph{Evidence injection.} GRNStyle injects the source/mask residual at the input of the 1st, 13th, and 25th transformer blocks, covering early, middle, and late stages; no residual is injected at the other block boundaries. This conditioning path adds 8.50M parameters ($\sim$0.43\% of the 2B backbone).

\begin{figure*}[t]
    \centering

    \begin{minipage}[t]{0.49\linewidth}
        \centering
        \includegraphics[width=\linewidth]{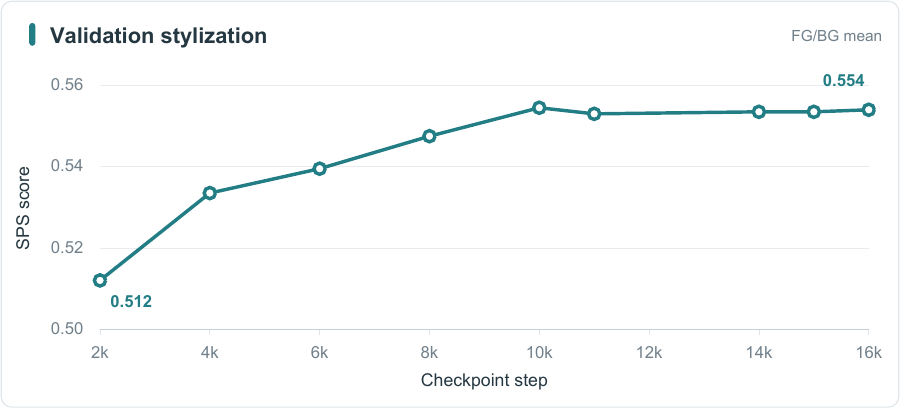}
    \caption{Validation stylization quality measured by mean foreground/background SPS.}
    \label{fig:grnstyle_validation_stylization}
    \end{minipage}%
    \hfill
    \begin{minipage}[t]{0.49\linewidth}
        \centering
        \includegraphics[width=\linewidth]{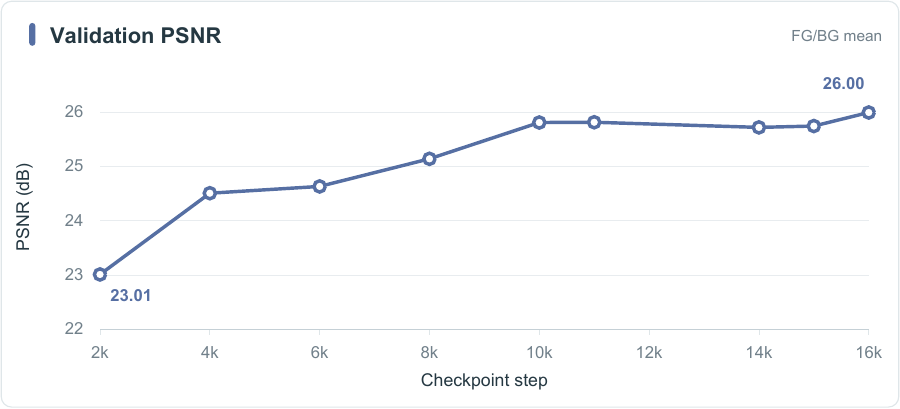}
    \caption{Validation preservation quality measured by mean foreground/background PSNR on non-target regions.}
    \label{fig:grnstyle_validation_psnr}
    \end{minipage}

\end{figure*}

We monitor local-control quality with two region-wise measures: SPS on the
edited region and PSNR on the preserved region. At each checkpoint, we
compute foreground and background scores separately and combine them with
an unweighted arithmetic mean. We evaluate nine checkpoints during
training. These measures complement
full-video bit-prediction statistics, which aggregate edited and preserved
tokens and can therefore obscure localized failures.

Figures~\ref{fig:grnstyle_validation_stylization} and~\ref{fig:grnstyle_validation_psnr}
report SPS and PSNR on the validation set.
Markers denote the nine evaluated
checkpoints, and the connecting lines are not smoothed. From update 2,000 to
16,000, mean SPS increases from 0.512 to 0.554, and mean PSNR from 23.01 to
26.00\,dB. Both metrics change more rapidly before update 10,000 and more
slowly thereafter, suggesting diminishing returns for local-control quality.

\section{Inference Efficiency}
\label{app:efficiency}

\begin{table}
\centering
\caption{Inference cost of the two task-specific models in Table~\ref{tab:local_stylization_main}. Lower values are better for both metrics.}
\label{tab:inference_cost}
\small
\begin{tabular}{lcc}
\toprule
Model & Latency (s)$\downarrow$ & GPU memory (GB)$\downarrow$ \\
\midrule
VACE-1.3B (our FT) & 39.2 & \textbf{23.7} \\
GRNStyle-2B (ours) & \textbf{34.5} & 27.4 \\
\bottomrule
\end{tabular}
\end{table}

Table~\ref{tab:local_stylization_main}
in the main paper contains two
task-specific configurations developed in this work: the fine-tuned
VACE-1.3B and GRNStyle-2B. Table~\ref{tab:inference_cost} compares their
inference cost to characterize the practical deployment profile of our
approach.
GRNStyle-2B requires 34.5\,s per inference and 27.4\,GB of GPU memory, compared with
39.2\,s and 23.7\,GB for VACE-1.3B (our FT). Relative to this 1.3B
configuration, GRNStyle-2B reduces latency by 12.0\% while increasing memory
by 15.6\%; the additional 3.7\,GB therefore yields a 4.7\,s reduction in
inference time. These results indicate that competitive local stylization can be achieved
with a 2B backbone, without relying on a 14B foundation model. Both
configurations fit within a single-device memory footprint, and GRNStyle-2B
achieves lower latency than the fine-tuned 1.3B variant, supporting practical
deployment without a multi-device serving setup.

\section{Global Stylization Comparison}
\label{app:global_stylization}

\begin{table}[t]
\centering
\caption{\textbf{Global stylization comparison.} All methods use the same 80 benchmark videos and ten-style subset.
$^{\dagger}$EchoStyle supports five styles and is evaluated on those only.
Best and second-best results are shown in bold and underlined, respectively.}
\label{tab:global_stylization}
\small
\setlength{\tabcolsep}{3pt}
\begin{tabular}{lcccccc}
\toprule
Method & Params & Venue & CSD$\uparrow$ & CLIPT$\uparrow$ & TC$\uparrow$ & Style$\uparrow$ \\
\midrule
StyleMaster~\cite{ye2025stylemaster} & 1.3B & CVPR '25 & 0.428 & 0.189 & 0.962 & 2.12 \\
Ditto~\cite{bai2026scaling} & 14B & CVPR '26 & \underline{0.649} & \textbf{0.207} & 0.976 & \textbf{4.73} \\
EchoStyle$^{\dagger}$~\cite{li2026echostyle} & 14B & ECCV '26 & 0.406 & \underline{0.190} & \underline{0.977} & 4.14 \\
PickStyle~\cite{mehraban2025pickstyle} & 14B & ECCV '26 & 0.388 & 0.182 & \textbf{0.984} & 3.04 \\
\midrule
\textbf{GRNStyle-2B} & 2B & -- & \textbf{0.683} & \underline{0.190} & \underline{0.977} & \underline{4.67} \\
\bottomrule
\end{tabular}%
\end{table}

To test whether GRNStyle extends beyond local editing, we set the editing
mask to all ones and apply the same checkpoint globally, without retraining.
We compare with existing global video stylization methods (including StyleMaster~\cite{ye2025stylemaster}, Ditto~\cite{bai2026scaling}, EchoStyle~\cite{li2026echostyle}, and PickStyle~\cite{mehraban2025pickstyle}) on the 80
benchmark videos and the same ten-style subset used in the main comparison.
All methods follow the same frame-alignment and metric implementations.
EchoStyle supports only five of the ten styles and is therefore evaluated on
those five.

Table~\ref{tab:global_stylization} reports CSD, CLIPT, TC, and VLM-based Style score. GRNStyle obtains the highest CSD (0.683) and the
second-highest VLM Style score (4.67). Its CLIPT (0.190) matches EchoStyle and is
close to the best result (0.207), while its TC (0.977) remains competitive
with the strongest methods. Notably, GRNStyle outperforms the 14B methods on
CSD despite using a 2B backbone, indicating that our style-aware conditioning
transfers from local to global stylization without a separate global model.

\begin{table*}
\centering
\caption{\textbf{Mask perturbation sensitivity.} NONE denotes the unperturbed baseline.
All variants use the same source videos, prompts, seeds, and sampling configuration; only the input mask is perturbed.
Subscripts denote the signed change from NONE; changes are computed from unrounded values.}
\label{tab:mask_sensitivity}
\resizebox{\linewidth}{!}{%
\begin{tabular}{l|ccccccc}
\toprule
Perturbation
& SPS-fg$\uparrow$ & SPS-bg$\uparrow$
& CLIPT-fg$\uparrow$ & CLIPT-bg$\uparrow$
& PSNR-fg$\uparrow$ & PSNR-bg$\uparrow$
& TC$\uparrow$ \\
\midrule
NONE & 0.388 & 0.677 & 0.154 & 0.188 & 26.6 & 26.3 & 0.972 \\
\midrule
\multicolumn{8}{@{}l}{\textit{Light}} \\
Boundary & 0.388$_{\scriptstyle+0.000}$ & 0.676$_{\scriptstyle-0.001}$ & 0.154$_{\scriptstyle+0.000}$ & 0.187$_{\scriptstyle-0.001}$ & 26.5$_{\scriptstyle-0.05}$ & 26.0$_{\scriptstyle-0.27}$ & 0.972$_{\scriptstyle+0.000}$ \\
Dilation & 0.394$_{\scriptstyle+0.006}$ & 0.677$_{\scriptstyle+0.000}$ & 0.155$_{\scriptstyle+0.001}$ & 0.187$_{\scriptstyle-0.001}$ & 26.5$_{\scriptstyle-0.04}$ & 25.8$_{\scriptstyle-0.49}$ & 0.972$_{\scriptstyle+0.000}$ \\
Erosion & 0.382$_{\scriptstyle-0.006}$ & 0.676$_{\scriptstyle-0.001}$ & 0.154$_{\scriptstyle+0.000}$ & 0.187$_{\scriptstyle-0.001}$ & 26.6$_{\scriptstyle+0.06}$ & 26.5$_{\scriptstyle+0.29}$ & 0.972$_{\scriptstyle+0.000}$ \\
Temporal & 0.388$_{\scriptstyle+0.000}$ & 0.677$_{\scriptstyle+0.000}$ & 0.154$_{\scriptstyle+0.000}$ & 0.188$_{\scriptstyle+0.000}$ & 26.6$_{\scriptstyle-0.01}$ & 26.2$_{\scriptstyle-0.07}$ & 0.972$_{\scriptstyle+0.000}$ \\
Dropout & 0.382$_{\scriptstyle-0.006}$ & 0.655$_{\scriptstyle-0.022}$ & 0.154$_{\scriptstyle+0.000}$ & 0.183$_{\scriptstyle-0.005}$ & 26.6$_{\scriptstyle+0.03}$ & 26.3$_{\scriptstyle+0.05}$ & 0.972$_{\scriptstyle+0.000}$ \\
\midrule
\multicolumn{8}{@{}l}{\textit{Strong}} \\
Boundary & 0.388$_{\scriptstyle+0.000}$ & 0.676$_{\scriptstyle-0.001}$ & 0.154$_{\scriptstyle+0.000}$ & 0.188$_{\scriptstyle+0.000}$ & 26.5$_{\scriptstyle-0.05}$ & 25.9$_{\scriptstyle-0.34}$ & 0.972$_{\scriptstyle+0.000}$ \\
Dilation & 0.403$_{\scriptstyle+0.015}$ & 0.681$_{\scriptstyle+0.004}$ & 0.156$_{\scriptstyle+0.002}$ & 0.188$_{\scriptstyle+0.000}$ & 26.2$_{\scriptstyle-0.32}$ & 22.2$_{\scriptstyle-4.01}$ & 0.972$_{\scriptstyle+0.000}$ \\
Erosion & 0.365$_{\scriptstyle-0.023}$ & 0.675$_{\scriptstyle-0.002}$ & 0.153$_{\scriptstyle-0.001}$ & 0.188$_{\scriptstyle+0.000}$ & 26.8$_{\scriptstyle+0.27}$ & 26.8$_{\scriptstyle+0.59}$ & 0.972$_{\scriptstyle+0.000}$ \\
Temporal & 0.388$_{\scriptstyle+0.000}$ & 0.677$_{\scriptstyle+0.000}$ & 0.154$_{\scriptstyle+0.000}$ & 0.188$_{\scriptstyle+0.000}$ & 26.5$_{\scriptstyle-0.07}$ & 25.9$_{\scriptstyle-0.35}$ & 0.972$_{\scriptstyle+0.000}$ \\
Dropout & 0.368$_{\scriptstyle-0.020}$ & 0.599$_{\scriptstyle-0.078}$ & 0.152$_{\scriptstyle-0.002}$ & 0.176$_{\scriptstyle-0.012}$ & 26.7$_{\scriptstyle+0.16}$ & 26.4$_{\scriptstyle+0.13}$ & 0.972$_{\scriptstyle+0.000}$ \\\bottomrule
\end{tabular}%
}
\end{table*}

\section{Robustness to Imperfect Masks}
\label{app:mask_sensitivity}

We evaluate the robustness of GRNStyle to imperfect user-provided masks while
keeping the source videos, style prompts, random seeds, guidance scale, and
sampling configuration fixed. Starting from a 40-case subset of the benchmark and ten
styles, we obtain 400 video--style evaluations for each variant.

We consider five perturbation categories and evaluate each at light and strong
severity levels:
\begin{itemize}[leftmargin=1.2em,itemsep=1pt,topsep=2pt]
  \item \textbf{Boundary noise}: pixels within a six-pixel band around the mask
  contour are independently flipped with probability 0.15 (light) or 0.35
  (strong) per frame.
  \item \textbf{Dilation}: the white edit region is expanded by a fixed radius
  of 3 (light) or 9 (strong) pixels.
  \item \textbf{Erosion}: the white edit region is contracted by a fixed radius
  of 3 (light) or 9 (strong) pixels.
  \item \textbf{Temporal jitter}: every frame receives an independently sampled
  signed radius in $[-3,3]$ (light) or $[-9,9]$ (strong) pixels.
  \item \textbf{Frame dropout}: the entire mask is cleared in approximately 10\%
  (light) or 30\% (strong) of frames.
\end{itemize}
Radii are specified at a 480p short side and scaled to the actual input resolution.
The perturbed mask replaces the original mask in both the polar style encoding and the geometric support path; no retraining or parameter update is involved.

\cref{tab:mask_sensitivity} reports the results. Under light perturbations,
all metrics remain close to the unperturbed baseline: absolute changes are within
0.006 for SPS and CLIPT and within 0.49 dB for PSNR, while TC remains at 0.972.
Under strong perturbations, erosion and dropout produce the clearest reduction
in stylization quality. Strong erosion lowers foreground SPS from 0.388 to
0.365, and strong dropout reduces background SPS from 0.677 to 0.599 and
background CLIPT from 0.188 to 0.176. In contrast, strong dilation leaves the
style-related metrics stable or slightly higher, but decreases background PSNR
from 26.3 to 22.2 dB because the expanded edit region inevitably modifies
content outside the original preserved area; its remaining metrics change
little. Overall, GRNStyle remains robust to imperfect masks: style fidelity is
largely preserved, and the remaining degradation is concentrated in the metric
directly affected by each geometric perturbation.

\section{Limitations and Future Work}
\label{app:limitations}

Our current system has three notable limitations. First, temporal consistency can degrade for fast-moving objects, especially near motion boundaries. Rapid motion makes it difficult to maintain stable region alignment and coherent stylization across adjacent frames, which may lead to local flicker or inconsistent boundaries. Improving motion-aware conditioning and boundary tracking is an important direction for future work.

Second, the data-construction pipeline relies on an object-removal model that may also remove the object's cast shadow. As a result, some paired targets lack the original shadow even when it should remain semantically consistent with the scene. Training on such pairs can encourage the stylization model to erase shadows, producing visually unnatural results. Future work should improve shadow-aware data construction and incorporate explicit supervision for preserving scene-consistent shadows.

Third, our evaluation suite does not include a boundary-adherence metric, although boundary accuracy is an important aspect of local editing quality. Such a metric is difficult to define in practice: some styles intentionally expand the target object (e.g., Simpsons-style stylization) while others shrink it, so no single fixed boundary can serve as the ground truth for all outputs. Future work should develop style-aware boundary evaluation that distinguishes intended shape deformation from mask leakage and inadvertent boundary violations.

\section{Additional Qualitative Results}
\label{sec:additional_qualitative}

\begin{figure*}[t]
    \centering
    \includegraphics[width=0.94\linewidth,page=1]{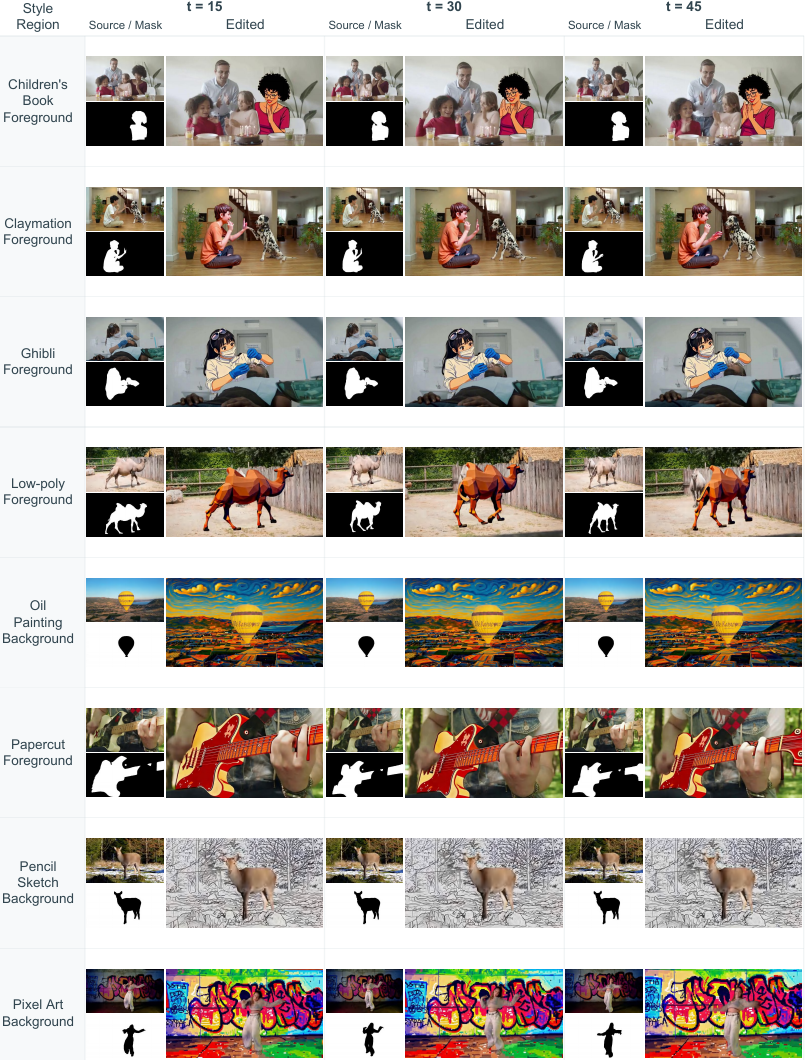}
    \caption{Additional qualitative results of GRNStyle. The 15th, 30th, and 45th frames of the generated videos are shown for each case.}
    \label{fig:grnstyle_results_1}
\end{figure*}

\begin{figure*}[t]
    \centering
    \includegraphics[width=0.94\linewidth,page=2]{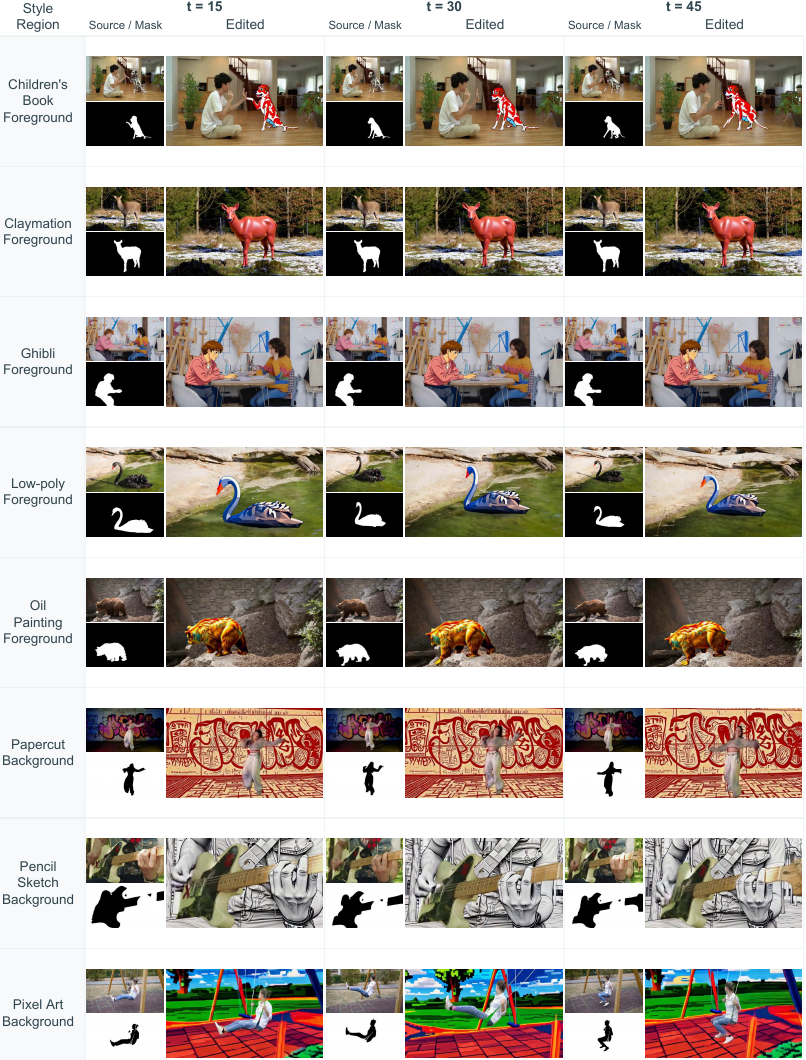}
    \caption{Additional qualitative results of GRNStyle. The 15th, 30th, and 45th frames of the generated videos are shown for each case.}
    \label{fig:grnstyle_results_2}
\end{figure*}

\begin{figure*}[t]
    \centering
    \includegraphics[width=0.94\linewidth,page=3]{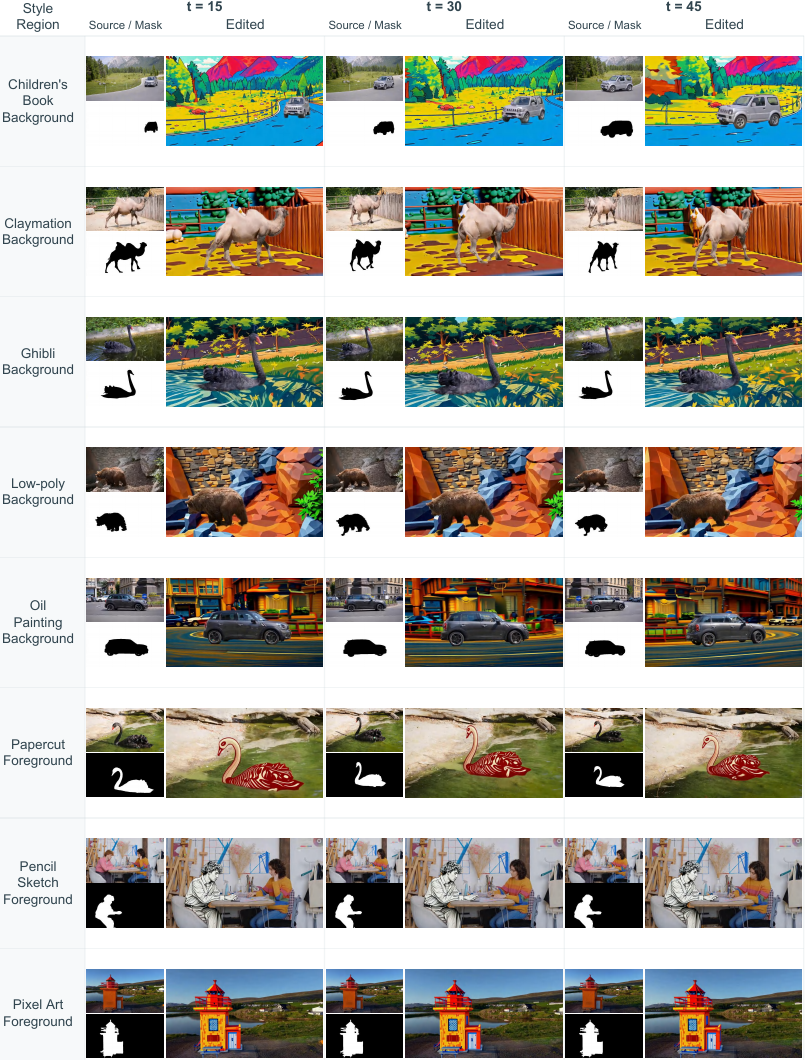}
    \caption{Additional qualitative results of GRNStyle. The 15th, 30th, and 45th frames of the generated videos are shown for each case.}
    \label{fig:grnstyle_results_3}
\end{figure*}

\begin{figure*}[t]
    \centering
    \includegraphics[width=0.94\linewidth,page=4]{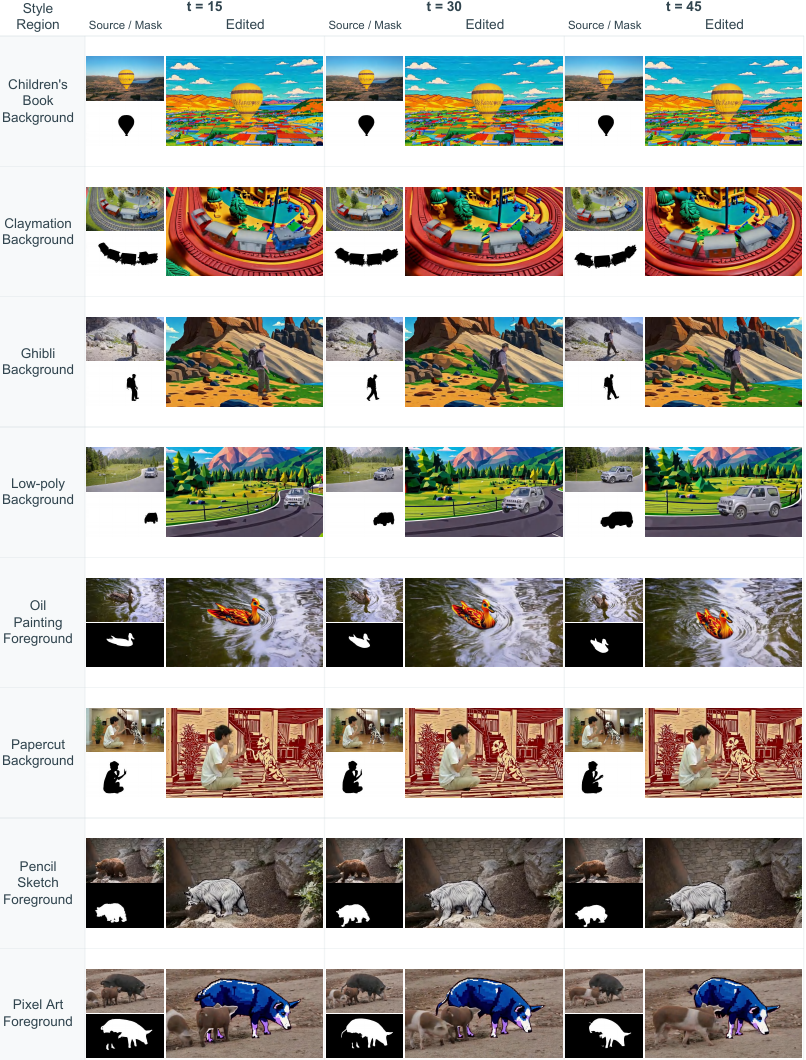}
    \caption{Additional qualitative results of GRNStyle. The 15th, 30th, and 45th frames of the generated videos are shown for each case.}
    \label{fig:grnstyle_results_4}
\end{figure*}

\begin{figure*}[t]
    \centering
    \includegraphics[width=0.94\linewidth,page=5]{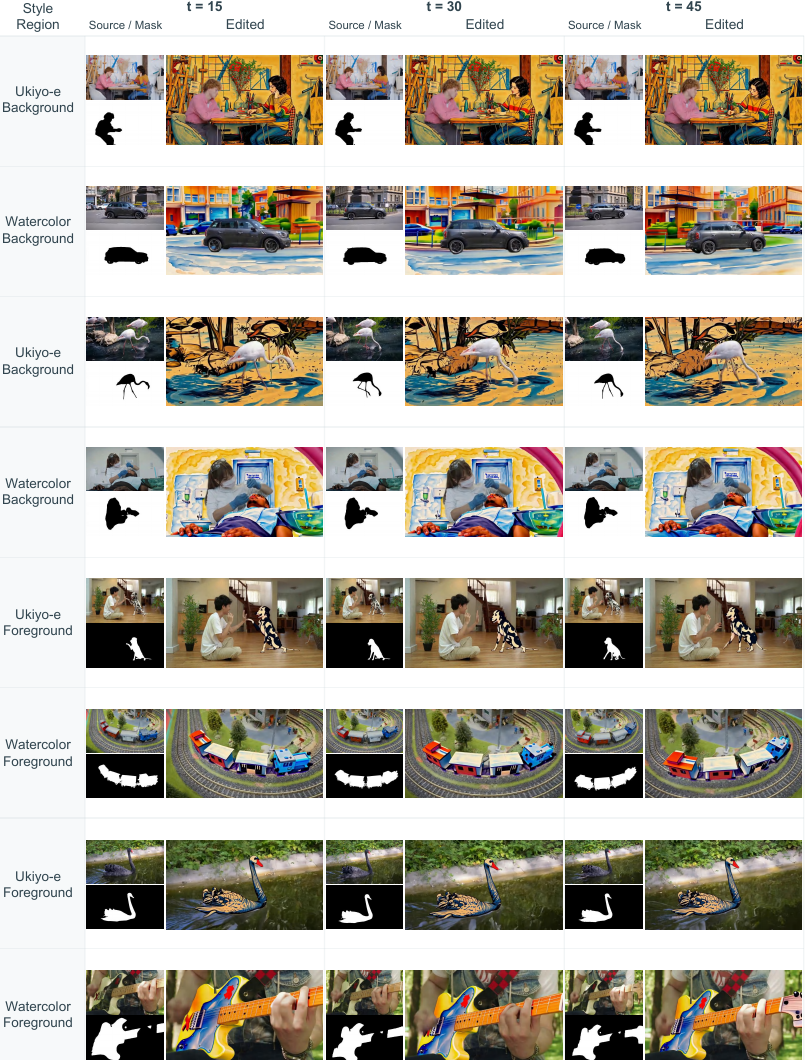}
    \caption{Additional qualitative results of GRNStyle. The 15th, 30th, and 45th frames of the generated videos are shown for each case.}
    \label{fig:grnstyle_results_5}
\end{figure*}

We apply GRNStyle to foreground and background regions in the LociArt Benchmark with diverse target styles. \cref{fig:grnstyle_results_1,fig:grnstyle_results_2,fig:grnstyle_results_3,fig:grnstyle_results_4,fig:grnstyle_results_5} show frames 15, 30, and 45 of each generated video, illustrating consistent regional stylization and preservation of surrounding content over time.

\end{document}